\documentclass{article}
\pdfoutput=1

\PassOptionsToPackage{numbers, compress}{natbib}
\usepackage[preprint]{neurips_2025}

\usepackage[utf8]{inputenc} % allow utf-8 input
\usepackage[T1]{fontenc}    % use 8-bit T1 fonts
\usepackage{hyperref}       % hyperlinks
\usepackage{url}            % simple URL typesetting
\usepackage{siunitx}
\usepackage{booktabs}       % professional-quality tables
\usepackage{amsfonts}       % blackboard math symbols
\usepackage{nicefrac}       % compact symbols for 1/2, etc.
\usepackage{microtype}      % microtypography
\usepackage[table]{xcolor}
\usepackage{graphicx}
\usepackage{comment}
\usepackage{multirow}
\usepackage{enumitem}
\usepackage[normalem]{ulem}
\newcommand{\eg}{{\textit{e.g.}}}
\newcommand{\ie}{{\textit{i.e.}}}

\definecolor{Light}{rgb}{0.99, 0.92, 0.95}
\definecolor{linkcolor}{RGB}{255,0,0}
\definecolor{urlcolor}{RGB}{255,105,180}
\definecolor{citecolor}{RGB}{66,168,235}
\definecolor{codegreen}{RGB}{57,182,74}
\hypersetup{colorlinks=true,linkcolor=linkcolor,urlcolor=urlcolor}

\newcommand{\dataset}{\textit{\textbf{M$^2$AD}}}
\newcommand{\datasetS}{\textit{\textbf{M$^2$AD-Synergy}}}
\newcommand{\datasetI}{\textit{\textbf{M$^2$AD-Invariant}}}

\title{Visual Anomaly Detection under Complex View-Illumination Interplay: \\ A Large-Scale Benchmark

}

\author{
\textbf{Yunkang Cao}$^{1,\dag}$, \textbf{Yuqi Cheng}$^{1,\dag}$, \textbf{Xiaohao Xu}$^{2}$, \textbf{Yiheng Zhang}$^{1}$,  \textbf{Yihan Sun}$^{1}$, \\
\textbf{Yuxiang Tan}$^{1}$, \textbf{Yuxin Zhang}$^{1}$, 
\textbf{Xiaonan Huang}$^{2}$,
\textbf{Weiming Shen}$^{1,}$\thanks{Corresponding Author, $^{\dag}$Contributed Equally.}
\\
$^{1}$State Key Laboratory of Intelligent Manufacturing Equipment and Technology, \\ Huazhong University of Science and Technology, Wuhan 430074, China \\ $^{2}$Robotics Department, University of Michigan, Ann Arbor, MI 48109 USA \\
\texttt{caoyunkang@ieee.org}, 
\texttt{yuqicheng@hust.edu.cn},
\texttt{xiaohaox@umich.edu}, \\
\texttt{yihengzhang@hust.edu.cn},
\texttt{yihansun@hust.edu.cn}, 
\texttt{yuxiangtan@hust.edu.cn},\\
\texttt{zyx\_hust@hust.edu.cn}, 
\texttt{xiaonanh@umich.edu},
\texttt{wshen@ieee.org}
}

\begin{document}

\maketitle

\begin{abstract}

The practical deployment of Visual Anomaly Detection (VAD) systems is hindered by their sensitivity to real-world imaging variations, particularly the complex interplay between viewpoint and illumination which drastically alters defect visibility. Current benchmarks largely overlook this critical challenge. We introduce \textit{\textbf{M}ulti-View \textbf{M}ulti-Illumination \textbf{A}nomaly \textbf{D}etection} (\dataset{}), a new large-scale benchmark comprising 119,880 high-resolution images designed explicitly to probe VAD robustness under such interacting conditions. By systematically capturing 999 specimens across 10 categories using 12 synchronized views and 10 illumination settings (120 configurations total), \dataset{} enables rigorous evaluation. We establish two evaluation protocols: \datasetS{} tests the ability to fuse information across diverse configurations, and \datasetI{} measures single-image robustness against realistic view-illumination effects. Our extensive benchmarking shows that state-of-the-art VAD methods struggle significantly on \dataset{}, demonstrating the profound challenge posed by view-illumination interplay. This benchmark serves as an essential tool for developing and validating VAD methods capable of overcoming real-world complexities. Our full dataset and test suite will be released at \href{hustcyq.github.io/M2AD}{\url{https://hustcyq.github.io/M2AD}} to facilitate the field.

% dataset homepage: https://hustcyq.github.io/M2AD

% benchmark code: https://github.com/hustCYQ/M2AD

\end{abstract}

\section{Introduction}

% \begin{figure}[h!]
% \centering\includegraphics[width=\linewidth]{figures/frame.pdf}
% \caption{\textcolor{blue}{\textbf{Why robust VAD demands multi-view, multi-illumination data.} \textbf{Left}: Real-world imaging complexities – viewpoint and lighting jointly determine if an anomaly is visible. \textbf{Right}: \dataset{} systematically captures these coupled variations across diverse materials to drive progress beyond idealized benchmarks.} \textcolor{gray}{Illustration for the influence of view, illumination, and materials --- highlighting the imperfect imaging nature}
% %左：数据集中的某个实例，分别用不同颜色的箭头表示相机、光源方向 -> 给示例图片，说明1. 遮挡问题/景深问题； 2. 光照和缺陷的敏感性；
% %要有工件的3D感，能否通过多视角多光照图像恢复3D结构？或者直接去重新拍摄+抠图
% %右：我们拍摄原理的示意：给出多视角多光照的示例图片，表示在某些configuration下，能够更好的成像
% }
% \vspace{-3mm}
% \label{fig:teaser}
% \end{figure}

Visual Anomaly Detection (VAD) is crucial for applications ranging from industrial quality control to medical imaging, aiming to identify deviations from normality. While benchmark datasets like MVTec AD~\cite{MVTec-AD}, VisA~\cite{VisA}, and Real-IAD~\cite{Real-IAD} have catalyzed significant algorithmic progress, a persistent gap remains between benchmark performance and reliable real-world deployment. We argue this gap stems fundamentally from the failure of existing benchmarks to capture the complexities of real-world imaging physics, particularly \textbf{the intricate interplay between viewpoint and illumination}.
%Visual anomaly detection (VAD) aims to identify abnormal instances and has seen substantial progress driven by benchmark datasets such as MVTec AD~\cite{MVTec-AD}, VisA~\cite{VisA}, and Real-IAD~\cite{Real-IAD}.  While these resources have propelled algorithmic development, existing state-of-the-art (SOTA) methods remain hindered by a critical performance gap when transitioning to practical deployment. This divergence originates from the underrepresentation of real-world imaging complexities in prevailing benchmarks.  

\begin{figure}[t]
\centering\includegraphics[width=\linewidth]{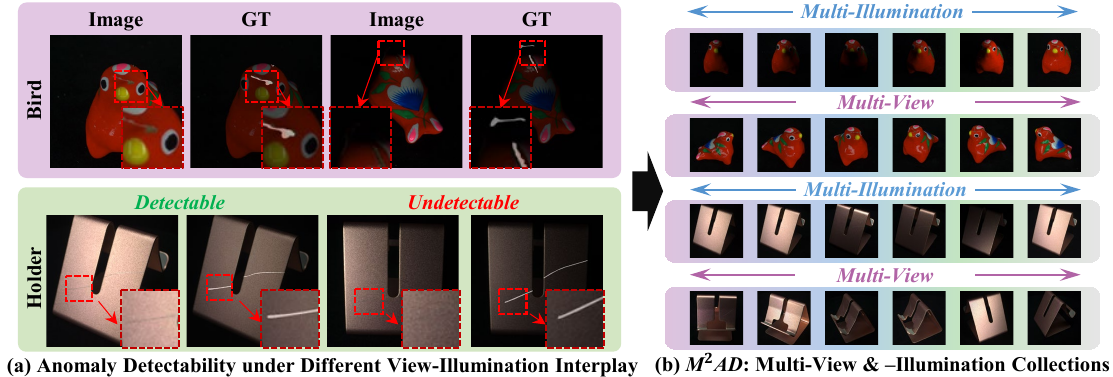}
\caption{
\textbf{Motivation.} \textbf{(a)} Anomaly detectability is governed by complex view-illumination interplay.  Each image pair shows the original input (left) alongside its corresponding ground truth (right), with anomaly regions highlighted in white.
\textbf{(b)} To address this challenge, our \dataset{} introduces multi-view and multi-illumination acquisition protocols, enabling robust anomaly detection across diverse conditions. Zoom in for a clearer view. More samples in \dataset{} are visualized in Appendix Sec.~\ref{supp:dataset_samples}.
}
\vspace{-3mm}
\label{fig:teaser}
\end{figure}

In practice, an object's visual appearance – and critically, the detectability of subtle anomalies like scratches or damages – is not static but a complex function of the geometric relationship between the camera, illumination sources, and the object's surface properties (see Fig.~\ref{fig:teaser}(a)). Factors like material reflectivity, surface curvature, and occlusions interact dynamically with viewing angle and lighting direction. A defect visible under direct lighting might vanish under diffuse light or from a different perspective. However, prevailing benchmarks often simplify reality, typically assuming near-ideal, constant imaging conditions or varying only \textbf{one} factor (view \textbf{or} illumination) in isolation~\cite{Real-IAD, PAD, MVTec_AD_2, Eyecandies}. This simplification prevents the evaluation of VAD methods against the compound challenges faced in realistic settings where view and illumination vary concurrently and interactively.

To bridge this critical evaluation gap, we introduce \textit{\textbf{M}ulti-View \textbf{M}ulti-Illumination \textbf{A}nomaly \textbf{D}etection} (\dataset{}) (Fig.~\ref{fig:teaser}(b)), the first large-scale VAD benchmark explicitly designed to model and evaluate robustness against complex view-illumination interplay under realistic conditions. \dataset{}'s core innovation lies in its systematic, synchronized capture methodology:
1)  \textbf{Controlled Interplay.} Each specimen is captured under 120 distinct, calibrated configurations resulting from the combination of 12 viewpoints and 10 illumination conditions, enabling fine-grained analysis of their joint effects. 2) \textbf{Scale and Diversity.} It comprises 119,880 images (69,070 normal, 50,810 anomalous) covering 999 unique specimens across 10 object categories with diverse materials (clay, plastic, wood, fabric, metal). 3) \textbf{High Fidelity.} Ultra-high resolution capture (3,648$\times$5,472 pixels) preserves sub-millimeter details crucial for detecting minute defects often masked by view-illumination effects.
4) \textbf{Generalization Challenge.} Each category includes two distinct sub-categories (\eg, differing color/size), providing a realistic testbed for generalizable VAD~\cite{AdaCLIP}.

%\dataset{} establishes new benchmarks through three core innovations. First, it introduces systematic variation across 120 distinct imaging configurations captured for each specimen, derived from the combination of 12 views and 10 illumination conditions. This coupled data allows for analysis of how view and illumination jointly influence anomaly appearance and detectability. Second, it embodies extensive categorical diversity, encompassing 10 object categories that feature diverse materials—including clay, plastic, wood, fabric, and metal—with 1,000 unique specimens. Third, it enables ultra-high-resolution image capture at 3,648$\times$5,472 pixels, thereby facilitating the detection of sub-millimeter anomalies.  Comprising 119,880 images (69,070 normal and 50,810 anomalous), the dataset further subdivides each category into two sub-categories (\textit{e.g.}, differing in color or size) to promote research on generalizable VAD \cite{AdaCLIP}. 

Leveraging this rich dataset, we propose two complementary benchmark setups:
\textbf{(1) \datasetS{}:} Evaluates a method's ability to synthesize information and achieve robust detection by utilizing the full 120 view-illumination configurations for a specimen. This directly probes performance leveraging the interplay.
\textbf{(2) \datasetI{}:} Assesses single-image robustness using standard protocols, but on images inherently containing the noise and variability arising from specific, complex view-illumination conditions within our capture process.

Our comprehensive evaluation of SOTA unsupervised VAD methods on these benchmarks reveals significant performance degradation compared to established datasets. For instance, Dinomaly~\cite{dinomaly}, despite achieving 99.6\% AUROC on MVTec AD, scores only 81.3\%  on our more challenging \datasetI{} setup. This stark difference validates \dataset{}'s ability to surface the limitations of current methods when confronted with realistic view-illumination interplay and underscores the need for new algorithmic approaches.

To sum up, our main contributions are:

\begin{itemize}[leftmargin=.5cm, itemsep=0pt, topsep=2pt]
    \item We introduce \dataset{}, the first large-scale VAD dataset capturing synchronized multi-view and multi-illumination images under realistic conditions, covering 120 imaging configurations for 999 specimens, in total of 119,880 images.
    \item We propose the \datasetS{} and \datasetI{} benchmarks, providing complementary paradigms for evaluating VAD methods. Our experiments demonstrate the significant challenge posed by view-illumination interplay to current SOTA methods, highlighting key areas for future research (\eg, robust fusion).
    \item We release our configurable imaging prototype design, facilitating reproducible research and adaptable data acquisition for diverse VAD scenarios.
    % \item \textcolor{red}{Adding main insights from experiments later}
\end{itemize}
% (Sec.~\ref{sec:data_acquistion}, \supp{xxx})

\section{Related Work}\label{sec:related-work}
\noindent\textbf{Benchmarks for Visual Anomaly Detection.} The landscape of VAD benchmarks has progressed from early, application-specific datasets~\cite{ksdd2,btad,MTD} to comprehensive 2D and 3D benchmarks like MVTec AD/3D~\cite{MVTec-AD, mvtec3d}, VisA~\cite{VisA}, Real3D-AD~\cite{read3d}, and Real-IAD/D$^3$~\cite{Real-IAD,Real-IAD-D3}, which established standard evaluation practices but often under simplified conditions. Subsequent efforts aimed to bridge the benchmark-to-reality gap by enhancing realism, primarily through incorporating either multi-view acquisitions using synchronized cameras~\cite{PAD,Real-IAD,Zhou2024RADAD,MANTA} to better capture geometry, or multi-illumination conditions, whether synthetic~\cite{Eyecandies} or real but often scale-limited~\cite{AeBAD,MVTec_AD_2}, to model appearance variations. However, these advancements typically addressed view and illumination challenges in isolation. Addressing this critical limitation, our \dataset{} introduces the first large-scale benchmark featuring systematically synchronized multi-view (12 viewpoints) and multi-illumination (10 conditions) capture, yielding 120 distinct configurations per specimen. This unique, structured data enables rigorous evaluation of method robustness against the complex interplay of these compound variables and supports advanced analysis techniques like photometric stereo~\cite{ZhangLLS22} and multi-view stereo~\cite{Fu0OT22} by providing the necessary controlled input variations. While \dataset{} involves sequences of images per object, its focus on identifying structural and surface defects through controlled geometric and photometric exploitation fundamentally distinguishes it from video anomaly detection frameworks~\cite{Li2025TowardsVD,Zhang2024HolmesVAUTL} concerned with temporal or semantic irregularities.

\noindent\textbf{Standard Visual Anomaly Detection Methods.}
Driven by conventional benchmarks like MVTec AD~\cite{MVTec-AD} and VisA~\cite{VisA}, most VAD methods adopt unsupervised learning paradigms using only normal training samples. Three principal approaches have emerged: reconstruction-based methods~\cite{dinomaly,INP-Former,Mambaad}, knowledge distillation frameworks~\cite{CDO,gu_remembering_2023}, and embedding-based techniques~\cite{GLASS,yao_hierarchical_2024,PatchCore}. SOTA unsupervised methods now achieve near-ceiling performance (>99\% image-level AUROC) on MVTec AD, suggesting benchmark saturation.
Recently, some researchers have started to explore the potential of generalizable VAD, which aims to develop a single model for multiple categories. Some also include unseen ones, a concept known as zero-shot anomaly detection \cite{AdaCLIP,anomalyclip}. However, existing methods typically train models on an auxiliary dataset and then test them on completely different datasets. Despite the promising vision, their performance remains limited. \dataset{} offers two similar sub-categories per product type, a scenario that is common in real-world applications where new products with slightly different characteristics emerge. By training with similar types, we can potentially derive a directly deployable model for new sub-categories, thus providing a new benchmark for generalizable VAD.

\noindent\textbf{Multi-Modal Visual Anomaly Detection Methods.}
Several studies have investigated multimodal inputs for enhanced anomaly detection. RGBD fusion approaches like M3DM~\cite{M3DM} and EasyNet~\cite{Easynet} demonstrate improved performance through deep feature fusion, while MulSen-AD~\cite{MulSen-AD} further extends modality integration to infrared imaging. Specialized methods have also emerged for multi-illumination~\cite{liu2024learning,zhang2024attention} and multi-view~\cite{he2024learning,kruse2024splatpose} analysis, yet no existing approach simultaneously addresses both modalities. The complex interaction of \dataset{}'s 120 configuration states further presents novel challenges in multimodal fusion and robustness optimization. We anticipate this benchmark will catalyze development of innovative methods capable of handling real-world multi-factor variations through adaptive feature composition and cross-modal reasoning.

\section{\textit{\textbf{M}ulti-View Multi-Illumination Anomaly Detection} (\textit{\textbf{\texorpdfstring{M$^2$AD}{M2AD}}}) Dataset}\label{sec:Dataset}

\begin{figure*}[ht]
\centering\includegraphics[width=\linewidth]{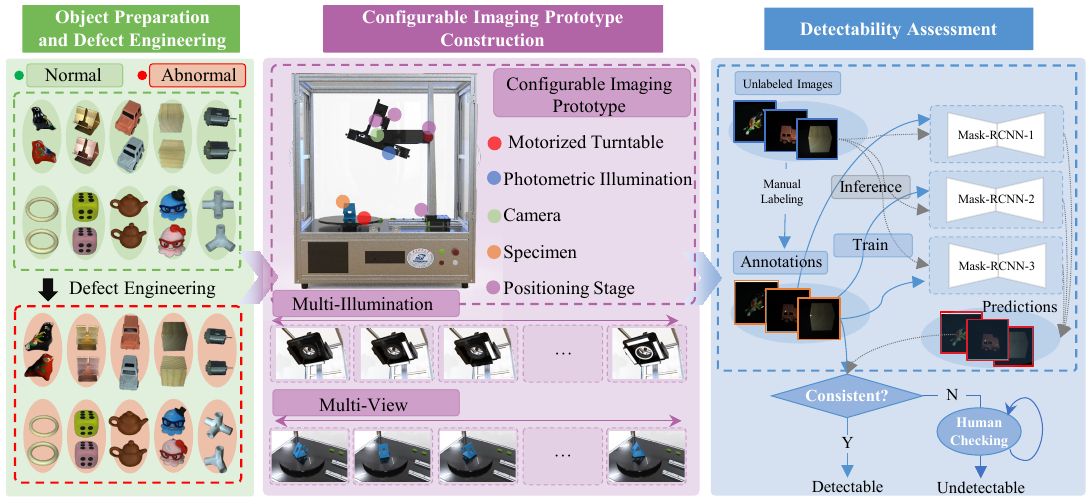}
\caption{\textbf{Data collection pipeline of \dataset{}.} A three-step process is employed. \textbf{(a)} Object preparation and defect engineering. \textbf{(b)} Design and construction of a configurable imaging prototype capable of capturing multi-view, multi-illumination images programmatically. \textbf{(c)} Assessing detectability by evaluating the consistency between predictions and annotations for \datasetI{}.}
% \vspace{-3mm}
\label{fig:dataset_construction}
\end{figure*}

\begin{table}[t]
\centering
\caption{{Statistical comparisons} between \dataset{} and existing 2D VAD datasets. \textbf{Our \dataset{} dataset is the first to include both multi-view and multi-illumination conditions.} \#Category, \#Image, and \#Configuration represent the number of categories, images, and imaging configurations, respectively.}
\label{tab:statistic}
\resizebox{1.\linewidth}{!}{
% Updated column specification: l|ccc|ccc|c|ccc
\begin{tabular}{l|ccc|ccc|c|ccc} 
\toprule[1.5pt]
\multirow{2}{*}{\textbf{Dataset}} &
  \multicolumn{3}{c|}{\#\textbf{Category}} &
  \multicolumn{3}{c|}{\#\textbf{Image}} &
  \multirow{2}{*}{\textbf{Image} \textbf{Resolution}} & % Moved Image Resolution Header
  \multicolumn{3}{c}{\#\textbf{Configuration}} \\ \cmidrule(lr){2-4} \cmidrule(lr){5-7} \cmidrule(lr){9-11} % Adjusted last cmidrule range
 &
  \textit{Main} &
  \textit{Sub.} &
  \textit{Total} &
  \textit{Normal} &
  \textit{Abnormal} &
  \textit{Total} &
   & % Placeholder for multirow Image Resolution
  \textit{View} &
  \textit{Illum.} &
  \textit{Total} \\ \midrule
MVTec AD~\cite{MVTec-AD} &
  15 &
  1 &
  15 &
  4,096 &
  1,258 &
  5,354 &
  700$\sim$1,024 & % Moved data
  1 &
  1 &
  1 \\ % Moved data
VisA~\cite{VisA} &
  12 &
  1 &
  12 &
  9,621 &
  1,200 &
  10,821 &
  960$\sim$1,562 & % Moved data
  1 &
  1 &
  1 \\ % Moved data
Real-IAD~\cite{Real-IAD} &
  30 &
  1 &
  30 &
  99,721 &
  51,329 &
  151,050 &
  2,000$\sim$5,000 & % Moved data
  5 &
  1 &
  5 \\ % Moved data
Eyecandies~\cite{Eyecandies} &
  10 &
  1 &
  10 &
  13,250 &
  2,250 &
  15,500 &
  512$\sim$512 & % Moved data
  1 &
  6 &
  6 \\ % Moved data
MANTA~\cite{MANTA} &
  5 &
  $\sim$8 &
  38 &
  652,455 &
  34,235 &
  686,690 &
  1016$\sim$1272 & % Moved data
  5 &
  1 &
  5 \\ % Moved data
PAD~\cite{PAD} &
  20 &
  1 &
  30 & % Note: Original had 30, but description says 20 main, 1 sub. Assuming 30 total is correct.
  5,231 &
  4,902 &
  10,133 &
  800$\sim$800 & % Moved data
  20 &
  1 &
  20 \\ % Moved data
MVTec AD 2~\cite{MVTec_AD_2} &
  8 &
  1 &
  8 &
  4,705 &
  3,299 &
  8,004 &
  1056$\sim$4224 & % Moved data
  1 &
  4 &
  4 \\ % Moved data
RAD~\cite{Zhou2024RADAD} &
  13 &
  1 &
  13 &
  2,535 &
  2,230 &
  4,765 &
  720$\sim$1280 & % Moved data
  68 &
  1 &
  68 \\ % Moved data
\midrule
\cellcolor{Light}\dataset{} (\textbf{Ours}) &
  \cellcolor{Light}10 &
  \cellcolor{Light}2 &
  \cellcolor{Light}20 &
  \cellcolor{Light}69,070 &
  \cellcolor{Light}50,810 &
  \cellcolor{Light}119,880 &
  \cellcolor{Light}3648$\sim$5472 & % Moved data
  \cellcolor{Light}12 &
  \cellcolor{Light}10 &
  \cellcolor{Light}120 \\ \bottomrule[1.5pt] % Moved data
\end{tabular}
}
\end{table}
\begin{figure*}[t]
\centering\includegraphics[width=\linewidth]{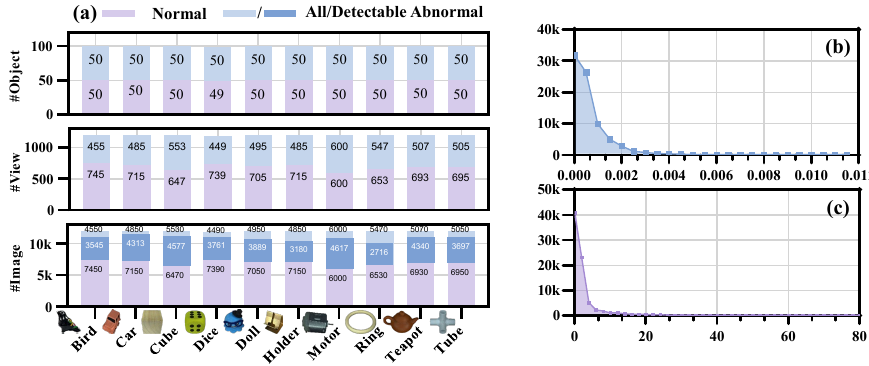}
\caption{\textbf{Statistics of \dataset{}}. \textbf{(a)} Distribution of normal and abnormal object, view, and image counts across different categories. ``Detectable'' refers to the abnormal images retained in Sec.~\ref{sec:data_acquistion}. \textbf{(b)} Percentage of image area occupied by anomaly regions. \textbf{(c)} Aspect ratio statistics of the minimum bounding rectangle of defects. }
% \vspace{-3mm}
\label{fig:dataset_statistics}
\end{figure*}

%\textcolor{red}{also count the number of undetectable images; the color for the figure a is a littble bit ugly, I would suggest to use lighter red and green to better align with the color style of figure b and c; for the figure b and c the number of y-axis can be changed into 10k, 20k, ..., 50k to make it more condensed} 

\subsection{Data Collection and Construction}~\label{sec:data_acquistion}
The \dataset{} construction pipeline, illustrated in Fig.~\ref{fig:dataset_construction}, follows a systematic three-stage methodology:

\noindent\textbf{1) Object Preparation and Defect Engineering.}
A diverse corpus of 20 physical objects was curated, organized into 10 main categories with dual sub-categories each, featuring diverse materials including clay, plastic, wood, fabric, and metallic compositions (more details are in Appendix Sec.~\ref{supp:object_selection}).  Representative specimens are shown in the upper segment of Fig.~\ref{fig:dataset_construction}(a). Diverse defect types was introduced, including perforations, surface abrasions, structural deformations, and bending anomalies, as shown in the lower panel of Fig.~\ref{fig:dataset_construction}(a). These engineered specimens, comprising both pristine and defective variants, were subsequently subjected by our configurable imaging prototype.

\noindent\textbf{2)  Configurable Imaging Prototype Construction.}
The proposed imaging prototype (Fig.~\ref{fig:dataset_construction}(b)) features an integrated modular architecture that synergistically combines programmable photometric illumination with precision angular positioning for comprehensive multi-modal image acquisition. A fixturing system maintains consistent specimen alignment relative to both illumination sources and imaging optics, ensuring geometric fidelity throughout acquisition cycles. Angular sampling is accomplished via a high-precision motorized turntable ($\pm\SI{0.5}{\degree}$ repeatability) acquiring twelve discrete specimen views through $\SI{30}{\degree}$ rotational increments. The photometric illumination module incorporates four linear bar lights and one coaxial ring light, operable independently or in synchronized combinations to generate ten distinct illumination regimes through programmable logic control. More details about the illumination setup and the collection process are in Appendix Sec.~\ref{supp:collection_details}. 

This architecture provides distinct advantages over conventional multi-camera systems like Real-IAD~\cite{Real-IAD} and MANTA~\cite{MANTA}. Our prototype constitutes the first implementation enabling concurrent variation of angular perspective and illumination conditions within a unified framework. Strategic integration of off-the-shelf components achieves cost efficiency: a single iRAYPLE A3B00CG000 industrial camera (resolution 3,648 $\times$ 5,472 pixels) interfaces with five programmable illumination sources and a single-axis rotational stage. This minimalist configuration yields exponential growth in imaging configuration diversity while maintaining hardware parsimony.

Deliberately eschewing mechanical complexity inherent in industrial inspection systems requiring specimen flipping, our design prioritizes methodological generality. Systematic photometric and angular sampling ensures comprehensive coverage of all exposed surfaces, with only occluded basal regions remaining unobserved. To enhance geometric adaptability, the optical assembly (illumination sources and camera) mounts on a four-degree-of-freedom positioning stage enabling translational and rotational adjustments relative to specimen morphology.

\noindent\textbf{3) Detectability Assessment.}
Following image acquisition, all anomalies were manually annotated and cross-verified to ensure labeling consistency. We derive two benchmarks, \ie, \datasetS{} and \datasetI{}. Recognizing substantial variations in anomaly detectability across view-illumination configurations -- where certain defects remain visually discernible only under specific acquisition parameters -- we formulated a systematic detectability assessment framework (Fig.~\ref{fig:dataset_construction}(c)). This methodology selectively retains only those image samples containing reliably detectable anomalies for inclusion in \datasetI{}. To operationalize this criterion, the dataset was partitioned into three mutually exclusive subsets, each employed to independently train Mask R-CNN detection architectures employing ResNeXt101 backbone networks. Quantitative discrepancies between model predictions and human annotations were systematically evaluated through two complementary metrics: intersection-over-union (IoU) spatial correspondence and prediction confidence scores. Samples exhibiting insufficient model-annotation alignment (IoU < 0.3) or low confidence predictions ($p$ < 0.5) were rigorously excluded. This systematic curation process guarantees \datasetI{} assessments focus exclusively on anomalies with reliable detection consensus.

\subsection{\textit{\textbf{\texorpdfstring{M$^2$AD}{M2AD}}} Dissection}
This section comprehensively analyzes \dataset{} on its characteristics and comparative advantages. 

\noindent\textbf{1) Comparative Dataset Analysis.}
Table~\ref{tab:statistic} presents statistical comparisons between \dataset{} and existing benchmarks. Our dataset distinguishes itself through three key innovations: First, it pioneers the integration of multi-illumination and multi-view configurations within a unified framework, encompassing 120 distinct imaging configurations. Second, it surpasses comparable datasets in spatial resolution. Third, with an extensive collection exceeding 100,000 images, \dataset{} rivals the scale of leading datasets like Real-IAD~\cite{Real-IAD} and MANTA~\cite{MANTA}.

\noindent\textbf{2) Data Statistics.}
Fig.~\ref{fig:dataset_statistics} presents statistical analysis of \dataset. Our dataset ensures balanced representation of normal and abnormal samples (Fig.~\ref{fig:dataset_statistics}(a)). {Also, we can see that only on average about 75\% of abnormal images are discerned as detectable, which are retained for \datasetI{}.}
Compared to existing datasets, our \dataset{} exhibits a smaller defect area proportion (Fig.~\ref{fig:dataset_statistics}(b)) and broader defect ratio range (Fig.~\ref{fig:dataset_statistics}(c)), indicating greater dataset complexity. This complexity is corroborated by the experimental results in Table~\ref{tab:M2AD_synergy} and Table~\ref{tab:M2AD_invariant}.

\noindent\textbf{3) Challenges and Prospects.}
Our dataset introduces three distinctive challenges that differentiate it from conventional VAD benchmarks. \textbf{Firstly}, the \textbf{\textit{enhanced heterogeneity}} stems from each category containing dual sub-categories captured under 120 distinct imaging configurations. This design induces substantial variation in normal specimen appearances, contrasting sharply with conventional datasets where normal samples maintain visual consistency across acquisition parameters. \textbf{Secondly}, the \textbf{\textit{subtle anomaly characteristics}} present unique detection challenges: carefully engineered anomalies may occupy merely 0.05\% of specimen volume or manifest as merely 4-pixel regions in $256\times256$ images, dimensions that approach the resolution limits of standard analytical methods. \textbf{Thirdly}, the \textbf{\textit{complex view-illumination interplay}} demands sophisticated interpretation. While our multi-configuration imaging protocol (120 variations) enables comprehensive specimen characterization, it simultaneously introduces configuration-dependent anomaly visibility—critical defects may only manifest under specific parameter combinations (Fig.~\ref{fig:teaser}(a)). Conversely, suboptimal configurations may introduce confounding artifacts such as specular reflections or low-signal regions. This inherent complexity necessitates holistic understanding of all imaging parameters for reliable anomaly identification. 
By closely approximating real-world operational conditions through these designed challenges, our dataset provides a more rigorous evaluation platform for VAD systems. We anticipate this resource will catalyze development of sophisticated analytical methods that explicitly address the intricate relationships between imaging physics and anomaly detection in practical implementations.

% Our dataset presents several key challenges compared to existing VAD benchmarks: First, it exhibits \textbf{\textit{higher heterogeneity}}. Each category in \dataset{} contains two sub-categories across 120 diverse imaging configurations, introducing significant variability in normal appearances compared to conventional datasets where normal data remains largely consistent. Second, it features \textbf{\textit{tiny anomalies}} designed to mimic real-world complexities. For instance, anomalies may occupy merely 0.05\% of the specimen volume, account for only about 4 pixels in a $256\times256$ image, almost imperceptible to conventional methods. Third, it demands comprehensive understanding of \textbf{\textit{view-illumination interplay}}. While our 120 configurations provide thorough specimen information, they present distinct challenges. Anomalies may only be detectable in specific configurations (Fig.~\ref{fig:teaser}(a)), while unsuitable setups might introduce noise such as reflective or overly dark regions. Mastery of all configurations is thus essential. Overall, our dataset better aligns with real-world applications. We anticipate it will inspire advanced techniques that fully account for imaging complexities in practical deployments.

% \usepackage{booktabs}
\begin{table}[t]
\centering
\caption{Benchmark results on \datasetS{} (listed as O-AUROC/I-AUROC/AUPRO) under the resolution of $256\times256$ ($224\times224$ for Dinomaly and INP-Former). Best results are in \textbf{bold}, and the second-best results are \uline{underlined}. 
}
\label{tab:M2AD_synergy}
\resizebox{\linewidth}{!}{
\setlength\tabcolsep{12.0pt}
\begin{tabular}{l|ccccc}
\toprule[1.5pt]
Category &
  \begin{tabular}[c]{@{}c@{}}CDO~\cite{CDO}\\      TII'2023\end{tabular} &
  \begin{tabular}[c]{@{}c@{}}RD++~\cite{RD++}\\      CVPR'2023\end{tabular} &
  \begin{tabular}[c]{@{}c@{}}MSFlow~\cite{zhou2024msflow}\\      TNNLS'2024\end{tabular} &
  \begin{tabular}[c]{@{}c@{}}Dinomaly~\cite{dinomaly}\\      CVPR'2025\end{tabular} &
  \begin{tabular}[c]{@{}c@{}}INP-Former~\cite{INP-Former}\\      CVPR'2025\end{tabular} \\ \midrule
Bird & 70.6/\uline{74.1}/\textbf{90.1} & \textbf{90.3}/70.2/79.8 & \uline{85.0}/62.0/71.4 & 75.1/\textbf{74.9}/\uline{86.9} & 80.0/67.2/84.1 \\
Car & 76.8/65.2/\uline{77.9} & \uline{85.0}/\uline{68.2}/75.6 & 67.9/55.9/67.4 & \textbf{86.7}/\textbf{75.1}/\textbf{78.3} & 58.1/53.9/72.1 \\
Cube & 72.2/64.9/72.4 & \textbf{83.1}/\uline{74.6}/\uline{80.7} & 66.0/57.8/58.7 & \uline{82.3}/\textbf{77.8}/\textbf{86.0} & 77.9/74.5/80.6 \\
Dice & 93.0/82.0/82.2 & \textbf{98.4}/\uline{89.4}/85.6 & 76.8/69.4/77.0 & \uline{98.1}/\textbf{93.0}/\uline{85.7} & 93.3/83.7/\textbf{87.7} \\
Doll & 69.9/64.0/74.4 & 66.8/65.9/85.4 & 56.4/55.1/68.9 & \textbf{74.4}/\uline{72.6}/\textbf{89.0} & \uline{72.5}/\textbf{73.7}/\uline{85.8} \\
Holder & 96.0/78.1/72.9 & 99.1/\textbf{87.8}/\uline{81.0} & 98.0/76.6/59.6 & \textbf{99.7}/\uline{85.8}/\textbf{90.0} & \uline{99.2}/76.4/\uline{81.0} \\
Motor & 83.7/69.7/94.0 & \uline{92.2}/\textbf{87.9}/\textbf{94.9} & 86.0/61.4/86.7 & \textbf{95.4}/\uline{85.4}/\uline{94.2} & 83.7/61.1/91.9 \\
Ring & \uline{91.6}/84.9/\uline{88.8} & \textbf{95.5}/\textbf{90.9}/77.2 & 74.7/72.4/83.9 & 91.2/\uline{87.3}/77.8 & 75.5/71.7/\textbf{91.4} \\
Teapot & \uline{92.6}/79.8/\uline{92.6} & 91.3/\uline{86.0}/91.7 & 83.0/63.9/77.3 & \textbf{99.9}/\textbf{94.6}/\textbf{94.3} & 91.6/79.1/92.4 \\
Tube & \uline{96.5}/\uline{81.8}/\textbf{93.7} & 92.1/81.2/\uline{90.9} & 89.0/67.3/84.1 & \textbf{97.2}/\textbf{83.3}/77.0 & 78.0/64.1/85.9 \\ \midrule
\cellcolor{Light}\textbf{Average} & \cellcolor{Light}84.3/74.4/83.9 & \cellcolor{Light}\uline{89.4}/\uline{80.2}/84.3 &\cellcolor{Light} 78.3/64.2/73.5 & \cellcolor{Light}\textbf{90.0}/\textbf{83.0}/\textbf{85.9} & \cellcolor{Light}81.0/70.5/\uline{85.3} \\ 
\bottomrule[1.5pt]
\end{tabular}
}
\end{table}
% \usepackage{booktabs}
\begin{table}[t]
\centering
\caption{Benchmark results on \datasetS{} (listed as O-AUROC/I-AUROC/AUPRO) under the resolution of $512\times512$ ($448\times448$ for Dinomaly and INP-Former). Best results are in \textbf{bold}, and the second-best results are \uline{underlined}. 
}
\label{tab:M2AD_synergy_512}
\resizebox{\linewidth}{!}{
\setlength\tabcolsep{12.0pt}
\begin{tabular}{l|ccccc}
\toprule[1.5pt]
Category &
  \begin{tabular}[c]{@{}c@{}}CDO~\cite{CDO}\\      TII'2023\end{tabular} &
  \begin{tabular}[c]{@{}c@{}}RD++~\cite{RD++}\\      CVPR'2023\end{tabular} &
  \begin{tabular}[c]{@{}c@{}}MSFlow~\cite{zhou2024msflow}\\      TNNLS'2024\end{tabular} &
  \begin{tabular}[c]{@{}c@{}}Dinomaly~\cite{dinomaly}\\      CVPR'2025\end{tabular} &
  \begin{tabular}[c]{@{}c@{}}INP-Former~\cite{INP-Former}\\      CVPR'2025\end{tabular} \\ \midrule
Bird & 73.8/\uline{71.8}/\uline{89.5} & \textbf{90.8}/71.3/79.9 & 86.8/61.8/78.8 & 86.8/\textbf{81.1}/\textbf{92.0} & \uline{87.7}/\uline{71.8}/89.2 \\
Car & 84.1/75.7/87.0 & \uline{86.3}/68.6/76.6 & 71.5/69.7/67.4 & \textbf{90.4}/\textbf{84.0}/\uline{90.2} & 85.6/\uline{80.6}/\textbf{91.9} \\
Cube & \uline{95.1}/\uline{91.1}/86.6 & 83.2/76.0/82.2 & 80.1/74.6/79.1 & \textbf{96.4}/\textbf{94.4}/\uline{92.7} & 89.4/86.5/\textbf{95.6} \\
Dice & 76.0/71.8/\textbf{93.7} & \textbf{98.5}/\textbf{90.0}/86.2 & \uline{79.1}/\uline{77.5}/91.6 & 71.9/72.6/\uline{92.1} & 72.5/74.4/90.9 \\
Doll & \uline{99.7}/\uline{90.4}/\uline{91.8} & 67.4/66.0/86.8 & 57.3/56.2/83.8 & \textbf{99.9}/\textbf{93.3}/\textbf{96.3} & 99.0/84.4/88.7 \\
Holder & 96.5/\uline{91.7}/93.8 & \textbf{99.1}/87.8/81.0 & 97.9/68.6/72.4 & \uline{98.8}/\textbf{94.3}/\textbf{98.2} & 80.5/70.0/\uline{96.2} \\
Motor & \uline{93.8}/\uline{90.5}/\textbf{98.8} & 92.2/87.9/94.9 & 77.3/61.7/93.4 & \textbf{95.9}/\textbf{95.3}/93.6 & 87.6/84.0/\uline{95.3} \\
Ring & 86.8/74.0/\textbf{96.1} & \textbf{96.7}/\textbf{91.6}/77.6 & 86.9/80.3/86.7 & \uline{94.9}/\uline{81.4}/\uline{91.6} & 69.8/60.8/80.6 \\
Teapot & \textbf{100.0}/\textbf{96.4}/\textbf{98.9} & 91.3/86.0/91.7 & 72.1/65.4/91.4 & \uline{98.9}/\uline{96.3}/\uline{98.6} & 82.1/80.2/96.7 \\
Tube & \textbf{95.9}/\uline{89.0}/\textbf{94.4} & 92.1/81.2/90.9 & 80.7/68.8/89.0 & \uline{95.7}/\textbf{90.7}/88.7 & 87.4/77.0/\uline{91.7} \\
\midrule
\cellcolor{Light}\textbf{Average} & \cellcolor{Light}\uline{90.2}/\uline{84.2}/\uline{93.0} & \cellcolor{Light}89.7/80.6/84.8 &\cellcolor{Light} 79.0/68.5/83.3 & \cellcolor{Light}\textbf{93.0}/\textbf{88.3}/\textbf{93.4} & \cellcolor{Light}84.2/77.0/91.7 \\ 
\bottomrule[1.5pt]
\end{tabular}
}
\end{table}

% \subsection{\dataset{} Visualization}

% \section{The Proposed Baseline}

% xxxxx xxxxx xxxxx xxxxx xxxxx xxxxx xxxxx xxxxx xxxxx xxxxx xxxxx xxxxx xxxxx xxxxx xxxxx xxxxx xxxxx xxxxx xxxxx xxxxx xxxxx xxxxx xxxxx xxxxx xxxxx xxxxx xxxxx xxxxx xxxxx xxxxx  xxxxx xxxxx xxxxx xxxxx xxxxx xxxxx xxxxx xxxxx xxxxx xxxxx xxxxx xxxxx xxxxx xxxxx xxxxx xxxxx xxxxx xxxxx xxxxx xxxxx xxxxx xxxxx xxxxx xxxxx xxxxx xxxxx xxxxx xxxxx xxxxx xxxxx  xxxxx xxxxx xxxxx xxxxx xxxxx xxxxx xxxxx xxxxx xxxxx xxxxx xxxxx xxxxx xxxxx xxxxx xxxxx xxxxx xxxxx xxxxx xxxxx

% \subsubsection{xxx}

\section{Benchmarking Results on \textit{\textbf{\texorpdfstring{M$^2$AD}{M2AD}}}}\label{sec:Benchmark}

\subsection{Benchmark Setups}
\textbf{1) \datasetS{} Setup.} This benchmark leverages the multi-view and multi-illumination configurations to evaluate VAD methods. Performance is assessed at two levels by aggregating predictions for each specimen. For object-level evaluation (O-AUROC), anomaly scores from all associated images are averaged. For view-level analysis, predictions from 10 spatially aligned images (same view, varying illumination) are aggregated to compute image-level AUROC (I-AUROC) and pixel-level AUPRO~\cite{MVTec-AD} for localization assessment.
%\textbf{\datasetS{} Setup.} This benchmark evaluates multi-modal visual data by leveraging the interplay between multi-illumination and multi-view configurations. Specifically, we aggregate predictions across all images of a single specimen to compute both object- and view-level performance metrics. For object-level evaluation, we average anomaly scores from all images and report the resulting object-level Area Under the Receiver Operating Characteristic curve (O-AUROC). For view-level analysis, we aggregate predictions across 10 spatially aligned images captured under varying illumination conditions. This enables the computation of image-level AUROC (I-AUROC) and pixel-level performance via the Area Under PRO curve (AUPRO)~\cite{MVTec-AD}, which evaluates localization accuracy.
% Additional metrics are detailed in \supp{A.2}.

\begin{figure*}[t]
\centering\includegraphics[width=\linewidth]{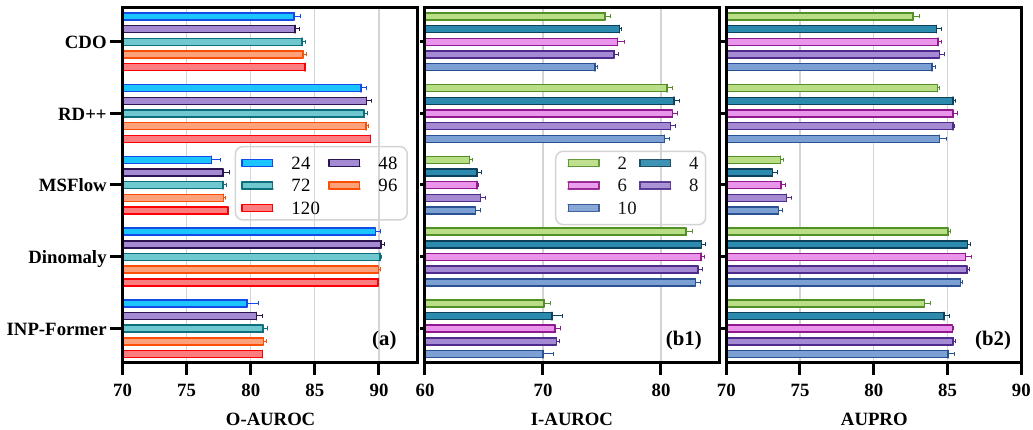}
\caption{\textbf{Ablation study results}.  
        \textbf{(a)} O-AUROC under different imaging configuration numbers (24, 48, 72, 96, and 120).
        \textbf{(b)} I-AUROC and AUPRO under different combinations of illumination conditions (2, 4, 6, 8, and 10). We randomly select the configurations and illumination conditions three times and report (mean $\pm$ std).}
% \vspace{-3mm}
\label{fig:ablation}
\end{figure*}

\textbf{2) \datasetI{} Setup.} This benchmark follows conventional methodologies (\eg, MVTec AD~\cite{MVTec-AD}) but incorporates additional imaging noise to better assess the robustness of VAD methods. Following standard practice, we evaluate performance using image-level AUROC (I-AUROC) and pixel-level AUPRO. Note that this evaluation only includes images deemed detectable (Sec.~\ref{sec:data_acquistion}).
%\textbf{\datasetI{} Setup.} This benchmark mirrors conventional methodologies like MVTec AD~\cite{MVTec-AD} but introduces more imaging noises, which can effectively evaluate the robustness of VAD methods. We adhere to the standard protocol by evaluating performance using image-level AUROC (I-AUROC) and pixel-level AUPRO.  Note that only images identified as detectable in Sec.~\ref{sec:data_acquistion} are evaluated in this setup.

\noindent \textbf{3) Benchmark Methods.}
We perform a comprehensive evaluation of representative SOTA approaches encompassing three dominant methodological paradigms: 
(i) \textit{knowledge distillation-based} methods, including CDO~\cite{CDO} and RD++~\cite{RD++};
(ii) \textit{embedding-based} approaches, as exemplified by MSFlow~\cite{zhou2024msflow}; and 
(iii) \textit{reconstruction-based} frameworks, comprising Dinomaly~\cite{dinomaly} and INP-Former~\cite{INP-Former}. 
To ensure a rigorous and reproducible evaluation, all experiments are conducted using official implementations with consistent parameter configurations. 
In alignment with conventional practices in the field, we adopt $256\times256$ resolution as the standard configuration for our experimental framework. For Dinomaly~\cite{dinomaly} and INP-Former~\cite{INP-Former}, we follow their original resolution of $224\times224$. All experiments are carried out on a single GeForce RTX 4090 GPU leveraging PyTorch 2.1.2.

% \textcolor{red}{if we follow 224x224 resolution, then what does the varied resolution settings in the ablation study mean?}. 
% Given that \dataset{} contains exceptionally minute anomalies that present detection challenges at conventional resolutions of $256\times256$, we adopt an enhanced input resolution of $512\times512$ as our baseline configuration.

% This adaptation is motivated by the empirical observation that increased spatial dimensions better preserve critical local texture details essential for identifying subtle anomalous patterns. 
% \supp{xx} further reveals a positive correlation between input resolution and detection performance, while simultaneously highlighting the non-trivial computational overhead associated with resolution scaling. 

% \subsection{Baseline: Uncertainty-guided Fusion}

% results \& in-depth analysis on illumination and views

\subsection{\textit{\textbf{\texorpdfstring{M$^2$AD-Synergy}{M2AD-Synergy}}} Results: Evaluating Multi-View/Multi-Illumination Synergy}

\noindent \textbf{1) Overall Performance.} Table~\ref{tab:M2AD_synergy} presents the main results on \datasetS{}. We observe a notable performance drop for all evaluated SOTA methods compared to their reported scores on benchmarks like MVTec AD. For instance, Dinomaly, often a top performer, achieves only 90.0\% O-AUROC and 83.0\% average I-AUROC here. This suggests that current methods, primarily designed for single-view, single-illumination data, struggle to effectively leverage or fuse information from the 120 diverse configurations provided in \datasetS{}. The complex interplay between viewpoint changes and varying illumination conditions poses a significant challenge not captured by previous benchmarks. However, performance varies across categories (\eg, Dinomaly reaches 99.9\% O-AUROC on `Teapot'), indicating that for certain object/defect types, the rich multi-configuration data can be highly informative even with existing methods. These results underscore the need for VAD models specifically designed for multimodal robustness and fusion.

\noindent \textbf{2) Impact of Input Resolution.} Many defects in \dataset{} are subtle. We investigate if standard low resolutions (256/224) hinder performance. Table~\ref{tab:M2AD_synergy_512} shows results using higher resolutions ($512\times512$ for 256-based methods, $448\times448$ for 224-based). Notably, CDO achieves a 5.8\% improvement in mean O-AUROC (84.3\% $\rightarrow$ 90.2\%), while Dinomaly shows a 3.0\% increase (90.0\% $\rightarrow$ 93.0\%). Particularly striking is the Dinomaly's O-AUROC on Cube improving from 82.6\% to 96.4\%, underscoring the resolution-dependent nature of fine defect detection. This aligns with findings in high-resolution VAD~\cite{VarAD} and suggests that standard resolutions may be insufficient for fine-grained industrial inspection tasks captured by \dataset{}. However, this improvement carries substantial computational cost (\eg, $\sim$4x memory for CNNs, $\sim$16x for ViTs when doubling resolution). This accuracy-efficiency trade-off, quantitatively characterized through our benchmark, underscores the need for novel architectural paradigms in high-resolution VAD. Future research directions should prioritize computationally sustainable frameworks that preserve \dataset{}'s intricate defect details while maintaining practical deployment feasibility—advancements that could significantly enhance real-world inspection systems' capacity to identify subtle anomalies in manufacturing environments.

% \usepackage{booktabs}
\begin{table}[t]
\centering
\caption{Benchmark results on \datasetI{} (listed as I-AUROC/AUPRO) under the resolution of $256\times256$ ($224\times224$ for Dinomaly and INP-Former). Best results are in \textbf{bold}, and the second-best results are \uline{underlined}. 
}
\label{tab:M2AD_invariant}
\resizebox{\linewidth}{!}{
\setlength\tabcolsep{12.0pt}
\begin{tabular}{l|ccccc}
\toprule[1.5pt]
Category &
  \begin{tabular}[c]{@{}c@{}}CDO~\cite{CDO}\\      TII'2023\end{tabular} &
  \begin{tabular}[c]{@{}c@{}}RD++~\cite{RD++}\\      CVPR'2023\end{tabular} &
  \begin{tabular}[c]{@{}c@{}}MSFlow~\cite{zhou2024msflow}\\      TNNLS'2024\end{tabular} &
  \begin{tabular}[c]{@{}c@{}}Dinomaly~\cite{dinomaly}\\      CVPR'2025\end{tabular} &
  \begin{tabular}[c]{@{}c@{}}INP-Former~\cite{INP-Former}\\      CVPR'2025\end{tabular} \\ \midrule
Bird & \uline{73.9}/\textbf{88.8} & 71.7/88.3 & 62.8/78.0 & \textbf{74.3}/\uline{88.6} & 69.1/85.6 \\
Car & 66.8/78.9 & \uline{70.9}/\uline{81.1} & 55.0/68.9 & \textbf{76.8}/\textbf{81.9} & 54.1/73.8 \\
Cube & 61.1/66.0 & \uline{71.0}/\uline{76.3} & 55.2/54.9 & \textbf{74.8}/\textbf{79.6} & 68.9/72.4 \\
Dice & 78.3/77.8 & \uline{87.8}/\textbf{83.1} & 66.0/72.8 & \textbf{89.7}/\uline{80.3} & 79.8/\uline{80.3} \\
Doll & 65.5/71.6 & 65.5/\uline{84.7} & 54.9/69.0 & \textbf{71.7}/\textbf{87.0} & \uline{69.8}/81.9 \\
Holder & 78.1/72.7 & \uline{87.3}/\uline{83.6} & 70.5/62.2 & \textbf{87.9}/\textbf{88.3} & 73.0/77.5 \\
Motor & 66.2/91.2 & \textbf{86.3}/\textbf{94.6} & 58.1/86.4 & \uline{84.9}/\uline{92.3} & 57.9/90.6 \\
Ring & 77.5/\uline{86.4} & \textbf{82.6}/71.7 & 65.3/81.9 & \uline{79.5}/69.3 & 65.9/\textbf{89.9} \\
Teapot & 75.0/88.7 & \uline{84.8}/\textbf{90.0} & 59.1/77.3 & \textbf{91.6}/89.6 & 74.3/\textbf{90.0} \\
Tube & 79.5/\textbf{93.4} & \uline{80.2}/\uline{91.0} & 57.9/82.7 & \textbf{81.5}/76.1 & 59.8/85.4 \\ \midrule
\cellcolor{Light}\textbf{Average} & \cellcolor{Light}72.2/81.6 & \cellcolor{Light}\uline{78.8}/\textbf{84.4} & \cellcolor{Light}60.5/73.4 & \cellcolor{Light}\textbf{81.3}/\uline{83.3} & \cellcolor{Light}67.3/82.7 \\ 
\bottomrule[1.5pt]
\end{tabular}
}
\end{table}

\noindent \textbf{3) Impact of Configuration Count.} We analyze how performance scales with the number of available configurations per specimen, sampling subsets (24, 48, 72, 96, 120) and evaluating O-AUROC (Fig.~\ref{fig:ablation}(a)). Contrary to intuition, using more configurations yields diminishing returns and can even degrade performance (\eg, RD++ drops 0.9\% O-AUROC from 48 to 72 configurations). A similar trend occurs when varying only the number of illuminations per view (Fig.~\ref{fig:ablation}(b)). This suggests that simple aggregation (averaging scores) struggles to effectively integrate information and may accumulate noise as more images are added, as evident in the prediction noises for multi-illumination images presented in Appendix Sec.~\ref{supp:am_multi_illumination}. It highlights limitations in current fusion strategies and points towards the need for more sophisticated approaches (\eg, feature-level fusion, attention mechanisms, selective view/illumination strategies) that can better exploit the rich information in \datasetS{} without being overwhelmed by redundancy or noise. Techniques inspired by photometric stereo or multi-view stereo could be promising future directions.

\subsection{\textit{\textbf{\texorpdfstring{M$^2$AD-Invariant}{M2AD-Invariant}}} Results: Evaluating Robustness to Imaging Noise}

\noindent \textbf{1) Quantitative Results.} Table~\ref{tab:M2AD_invariant} presents the performance summary on the \datasetI{} setup. Consistent with expectations, the incorporation of realistic imaging noise typically degrades performance relative to cleaner benchmarks such as MVTec AD. Specifically, the highest-performing method, Dinomaly~\cite{dinomaly}, achieved I-AUROC and AUPRO scores of 81.3\% and 83.3\%, respectively. Other methods obtained I-AUROCs below 80.0\%. These findings suggest that existing VAD methods are susceptible to environmental noise. Although such noise is present in the training set (all imaging configurations are utilized for training), current VAD methods may inadequately model it and fail to distinguish subtle anomalies from normality perturbed by extensive environmental noise. Future research is encouraged to either disentangle imaging noise from structural patterns, thereby focusing on detecting structural deviations, or to enhance VAD methods from low-level modeling of structural normality, which may vary across different environments, to high-level understanding of normality.

\begin{figure*}[t]
\centering\includegraphics[width=\linewidth]{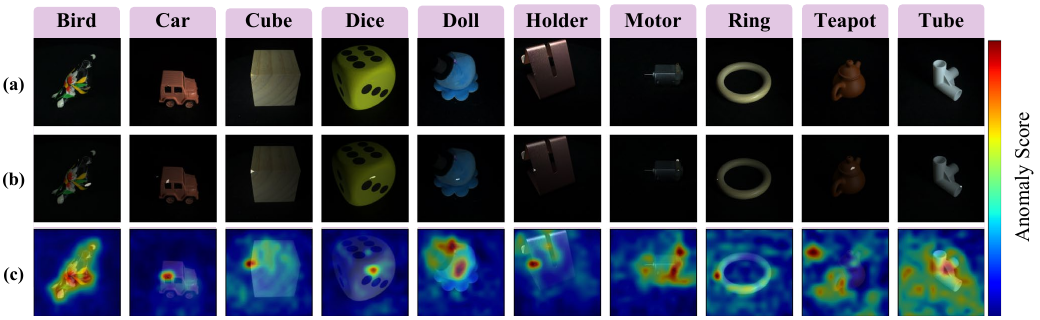}
\caption{\textbf{Visualization of anomaly detection results}. \textbf{(a)} Input image, \textbf{(b)} ground truth (anomalies highlighted in white), \textbf{(c)} predicted anomaly maps by the best-performing model Dinomaly~\cite{dinomaly}. Despite its robustness, the visualization demonstrates Dinomaly’s limitations in capturing anomalies across diverse scenarios. Zoom in for a clearer view. See Appendix Sec.~\ref{supp:anomaly_visualization} for more visualizations.
}
% \vspace{-3mm}
\label{fig:am_visualizations}
\end{figure*} % Consider placement

\noindent \textbf{2) Qualitative Analysis.} Fig.~\ref{fig:am_visualizations} presents qualitative results from the \datasetI{} evaluation. Consistent with their performance limitations in Table~\ref{tab:M2AD_invariant}, even SOTA methods such as Dinomaly~\cite{dinomaly} only identify certain anomalies (\eg, Car, Cube) while failing to detect more challenging cases (\eg, Doll, Motor). Notably, our \dataset{} explicitly incorporates subtle anomalies and environmental noise patterns characteristic of real-world scenarios. The observed performance gaps between existing VAD methods and our benchmark requirements underscore the necessity for more sophisticated anomaly detection frameworks capable of handling nuanced real-world variations.

\section{Conclusion}\label{sec:conclusion}
We introduced \dataset{}, the first large-scale VAD benchmark designed to address the critical challenge of view-illumination interplay, a major factor limiting the real-world deployment of current methods. By systematically capturing 120 synchronized view-illumination configurations for diverse objects, \dataset{} provides a unique resource for evaluating robustness against realistic imaging complexities. Our proposed \datasetS{} and \datasetI{} benchmarks revealed significant performance drops for SOTA methods compared to simpler datasets, confirming the difficulty posed by interacting view and illumination conditions and validating the need for such a benchmark.

\noindent\textbf{Limitations and Future Directions.} While \dataset{} represents significant progress in holistic VAD evaluation, the substantial data complexity (120 configurations per specimen) inherent in our design reveals several critical research avenues:

\begin{enumerate}[leftmargin=*,topsep=0pt]
    \item \textbf{Optimal Configuration Selection}: Developing principled methodologies for identifying minimal sufficient subsets of views and illuminations that preserve diagnostic information while maximizing acquisition efficiency. The controlled experimental setup of \dataset{} reduces this challenge to a tractable combinatorial optimization problem, where configuration subsets can be evaluated through our benchmark's structured validation protocol.
    
    \item \textbf{Multi-Modal Fusion Architectures}: Advancing beyond naive feature aggregation through novel fusion paradigms that explicitly model photometric-stereo relationships and geometric constraints. This includes attention-based feature disentanglement, physics-informed neural rendering, and cross-modal consistency learning -- directions particularly enabled by \dataset{}'s synchronized multi-view/multi-illumination structure.
    
    \item \textbf{Modality Contribution Analysis}: Leveraging \dataset{}'s factorial design to quantitatively decompose performance impacts of view diversity versus illumination variation, enabling data-driven optimization of inspection system configurations through saliency mapping and ablation studies.
    
    \item \textbf{Generalizable VAD Frameworks:} The dataset's dual sub-category organization supports development of zero-shot and few-shot VAD paradigms through cross-category transfer learning. This direction addresses a critical industrial need for anomaly detection systems that generalize across product lines without exhaustive retraining.
\end{enumerate}

\noindent Beyond these immediate directions, \dataset{} provides high-resolution imagery capturing subtle surface anomalies under controlled conditions -- a unique resource for developing high-precision VAD systems aligned with industrial inspection requirements. We anticipate this benchmark will catalyze progress toward view- and illumination-robust anomaly detection while bridging the gap between academic research and real-world industrial applications.

{\small
\bibliographystyle{unsrt}

% \bibliography{ref}

\begin{thebibliography}{10}

\bibitem{MVTec-AD}
Paul Bergmann, Kilian Batzner, Michael Fauser, David Sattlegger, and Carsten Steger.
\newblock The {MVTec} anomaly detection dataset: A comprehensive real-world dataset for unsupervised anomaly detection.
\newblock {\em International Journal of Computer Vision}, 129(4):1038--1059, 2021.

\bibitem{VisA}
Yang Zou, Jongheon Jeong, Latha Pemula, Dongqing Zhang, and Onkar Dabeer.
\newblock Spot-the-difference self-supervised pre-training for anomaly detection and segmentation.
\newblock In {\em European Conference on Computer Vision}, pages 392--408. Springer, 2022.

\bibitem{Real-IAD}
Chengjie Wang, Wenbing Zhu, Bin-Bin Gao, Zhenye Gan, Jiangning Zhang, Zhihao Gu, Shuguang Qian, Mingang Chen, and Lizhuang Ma.
\newblock Real-iad: A real-world multi-view dataset for benchmarking versatile industrial anomaly detection.
\newblock In {\em Proceedings of the IEEE/CVF Conference on Computer Vision and Pattern Recognition}, pages 22883--22892, 2024.

\bibitem{PAD}
Qiang Zhou, Weize Li, Lihan Jiang, Guoliang Wang, Guyue Zhou, Shanghang Zhang, and Hao Zhao.
\newblock Pad: A dataset and benchmark for pose-agnostic anomaly detection.
\newblock In A.~Oh, T.~Naumann, A.~Globerson, K.~Saenko, M.~Hardt, and S.~Levine, editors, {\em Advances in Neural Information Processing Systems}, volume~36, pages 44558--44571. Curran Associates, Inc., 2023.

\bibitem{MVTec_AD_2}
Lars Heckler-Kram, Jan-Hendrik Neudeck, Ulla Scheler, Rebecca K{\"o}nig, and Carsten Steger.
\newblock The mvtec ad 2 dataset: Advanced scenarios for unsupervised anomaly detection.
\newblock {\em ArXiv}, abs/2503.21622, 2025.

\bibitem{Eyecandies}
Luca Bonfiglioli, Marco Toschi, Davide Silvestri, Nicola Fioraio, and Daniele~De Gregorio.
\newblock The eyecandies dataset for unsupervised multimodal anomaly detection and localization.
\newblock In Lei Wang, Juergen Gall, Tat{-}Jun Chin, Imari Sato, and Rama Chellappa, editors, {\em Asian Conference on Computer Vision}, volume 13845 of {\em Lecture Notes in Computer Science}, pages 459--475. Springer, 2022.

\bibitem{AdaCLIP}
Yunkang Cao, Jiangning Zhang, Luca Frittoli, Yuqi Cheng, Weiming Shen, and Giacomo Boracchi.
\newblock Adaclip: Adapting clip with hybrid learnable prompts for zero-shot anomaly detection.
\newblock In {\em European Conference on Computer Vision}, 2024.

\bibitem{dinomaly}
Jia Guo, Shuai Lu, Weihang Zhang, Fang Chen, Huiqi Li, and Hongen Liao.
\newblock Dinomaly: The less is more philosophy in multi-class unsupervised anomaly detection.
\newblock {\em ArXiv}, 2025.

\bibitem{ksdd2}
Jakob Bo{\v{z}}i{\v{c}}, Domen Tabernik, and Danijel Sko{\v{c}}aj.
\newblock Mixed supervision for surface-defect detection: From weakly to fully supervised learning.
\newblock {\em Computers in Industry}, 129:103459, 2021.

\bibitem{btad}
Pankaj Mishra, Riccardo Verk, Daniele Fornasier, Claudio Piciarelli, and Gian~Luca Foresti.
\newblock {VT-ADL}: A vision transformer network for image anomaly detection and localization.
\newblock In {\em IEEE International Symposium on Industrial Electronics}, pages 01--06. IEEE, 2021.

\bibitem{MTD}
Yibin Huang, Congying Qiu, Yue Guo, Xiaonan Wang, and Kui Yuan.
\newblock Surface defect saliency of magnetic tile.
\newblock In {\em International Conference on Automation Science and Engineering (CASE)}, pages 612--617, 2018.

\bibitem{mvtec3d}
Paul Bergmann, Xin Jin, David Sattlegger, and Carsten Steger.
\newblock The mvtec 3d-ad dataset for unsupervised 3d anomaly detection and localization.
\newblock {\em Proceedings of the 17th International Joint Conference on Computer Vision, Imaging and Computer Graphics Theory and Applications}, 5:202--213, 2022.
\newblock doi: {10.5220/0010865000003124}.

\bibitem{read3d}
Jiaqi Liu, Guoyang Xie, Xinpeng Li, Jinbao Wang, Yong Liu, Chengjie Wang, Feng Zheng, et~al.
\newblock Real3d-ad: A dataset of point cloud anomaly detection.
\newblock In {\em Thirty-seventh Conference on Neural Information Processing Systems Datasets and Benchmarks Track}, volume~36, 2024.

\bibitem{Real-IAD-D3}
Wenbing Zhu, Lidong Wang, Ziqing Zhou, Chengjie Wang, Yurui Pan, Ruoyi Zhang, Zhuhao Chen, Linjie Cheng, Bin-Bin Gao, Jiangning Zhang, Zhenye Gan, Yuxie Wang, Yulong Chen, Shuguang Qian, Mingmin Chi, Bo~Peng, and Lizhuang Ma.
\newblock Real-iad d3: A real-world 2d/pseudo-3d/3d dataset for industrial anomaly detection.
\newblock 2025.

\bibitem{Zhou2024RADAD}
Kaichen Zhou, Yang Cao, Teawhan Kim, Hao Zhao, Hao Dong, Kai~Ming Ting, and Ye~Zhu.
\newblock Rad: A dataset and benchmark for real-life anomaly detection with robotic observations.
\newblock {\em ArXiv}, abs/2410.00713, 2024.

\bibitem{MANTA}
Lei Fan, Dongdong Fan, Zhiguang Hu, Yiwen Ding, Donglin Di, Kai Yi, Maurice Pagnucco, and Yang Song.
\newblock Manta: A large-scale multi-view and visual-text anomaly detection dataset for tiny objects.
\newblock {\em arXiv preprint arXiv:2412.04867}, 2024.

\bibitem{AeBAD}
Zilong Zhang, Zhibin Zhao, Xingwu Zhang, Chuang Sun, and Xuefeng Chen.
\newblock Industrial anomaly detection with domain shift: A real-world dataset and masked multi-scale reconstruction.
\newblock {\em Computers in Industry}, 151:103990, 2023.

\bibitem{ZhangLLS22}
Kai Zhang, Fujun Luan, Zhengqi Li, and Noah Snavely.
\newblock {IRON:} inverse rendering by optimizing neural sdfs and materials from photometric images.
\newblock In {\em {IEEE/CVF} Conference on Computer Vision and Pattern Recognition, {CVPR} 2022, New Orleans, LA, USA, June 18-24, 2022}, pages 5555--5564. {IEEE}, 2022.

\bibitem{Fu0OT22}
Qiancheng Fu, Qingshan Xu, Yew~Soon Ong, and Wenbing Tao.
\newblock Geo-neus: Geometry-consistent neural implicit surfaces learning for multi-view reconstruction.
\newblock In Sanmi Koyejo, S.~Mohamed, A.~Agarwal, Danielle Belgrave, K.~Cho, and A.~Oh, editors, {\em Advances in Neural Information Processing Systems 35: Annual Conference on Neural Information Processing Systems 2022, NeurIPS 2022, New Orleans, LA, USA, November 28 - December 9, 2022}, 2022.

\bibitem{Li2025TowardsVD}
Wenqiao Li, Yao Gu, Xintao Chen, Xiaohao Xu, Ming Hu, Xiaonan Huang, and Yingna Wu.
\newblock Towards visual discrimination and reasoning of real-world physical dynamics: Physics-grounded anomaly detection.
\newblock {\em ArXiv}, abs/2503.03562, 2025.

\bibitem{Zhang2024HolmesVAUTL}
Huaxin Zhang, Xiaohao Xu, Xiangdong Wang, Jia li~Zuo, Xiaonan Huang, Changxin Gao, Shanjun Zhang, Li~Yu, and Nong Sang.
\newblock Holmes-vau: Towards long-term video anomaly understanding at any granularity.
\newblock {\em ArXiv}, abs/2412.06171, 2024.

\bibitem{INP-Former}
Wei Luo, Yunkang Cao, Haiming Yao, Xiaotian Zhang, Jianan Lou, Yuqi Cheng, Weiming Shen, and Wenyong Yu.
\newblock Exploring intrinsic normal prototypes within a single image for universal anomaly detection.
\newblock {\em arXiv preprint arXiv:2503.02424}, 2025.

\bibitem{Mambaad}
Haoyang He, Yuhu Bai, Jiangning Zhang, Qingdong He, Hongxu Chen, Zhenye Gan, Chengjie Wang, Xiangtai Li, Guanzhong Tian, and Lei Xie.
\newblock {MambaAD}: {Exploring} {State} {Space} {Models} for {Multi}-class {Unsupervised} {Anomaly} {Detection}.
\newblock In Amir Globersons, Lester Mackey, Danielle Belgrave, Angela Fan, Ulrich Paquet, Jakub~M. Tomczak, and Cheng Zhang, editors, {\em Advances in {Neural} {Information} {Processing} {Systems}}, 2024.

\bibitem{CDO}
Yunkang Cao, Xiaohao Xu, Zhaoge Liu, and Weiming Shen.
\newblock Collaborative discrepancy optimization for reliable image anomaly localization.
\newblock {\em {IEEE} Transactions on Industrial Informatics}, pages 1--10, 2023.

\bibitem{gu_remembering_2023}
Zhihao Gu, Liang Liu, Xu~Chen, Ran Yi, Jiangning Zhang, Yabiao Wang, Chengjie Wang, Annan Shu, Guannan Jiang, and Lizhuang Ma.
\newblock Remembering {Normality}: {Memory}-guided {Knowledge} {Distillation} for {Unsupervised} {Anomaly} {Detection}.
\newblock pages 16355--16363, Paris, France, October 1-6, 2023, October 2023. IEEE.

\bibitem{GLASS}
Qiyu Chen, Huiyuan Luo, Chengkan Lv, and Zhengtao Zhang.
\newblock A unified anomaly synthesis strategy with gradient ascent for industrial anomaly detection and localization.
\newblock pages 37--54, Milan, Italy, September 29-October 4, 2024, 2025. Springer.

\bibitem{yao_hierarchical_2024}
Xincheng Yao, Ruoqi Li, Zefeng Qian, Lu~Wang, and Chongyang Zhang.
\newblock Hierarchical gaussian mixture normalizing flow modeling for unified anomaly detection.
\newblock In {\em European {Conference} on {Computer} {Vision}}, pages 92--108. Springer, 2024.

\bibitem{PatchCore}
Karsten Roth, Latha Pemula, Joaquin Zepeda, Bernhard Scholkopf, Thomas Brox, and Peter Gehler.
\newblock Towards {Total} {Recall} in {Industrial} {Anomaly} {Detection}.
\newblock pages 14298--14308, New Orleans, LA, USA, June 18-24, 2022, June 2022. IEEE.

\bibitem{anomalyclip}
Qihang Zhou, Guansong Pang, Yu~Tian, Shibo He, and Jiming Chen.
\newblock Anomalyclip: Object-agnostic prompt learning for zero-shot anomaly detection.
\newblock In {\em International Conference on Learning Representations,}, 2024.

\bibitem{M3DM}
Yue Wang, Jinlong Peng, Jiangning Zhang, Ran Yi, Yabiao Wang, and Chengjie Wang.
\newblock Multimodal industrial anomaly detection via hybrid fusion.
\newblock In {\em Proceedings of the IEEE/CVF Conference on Computer Vision and Pattern Recognition}, pages 8032--8041, 2023.

\bibitem{Easynet}
Ruitao Chen, Guoyang Xie, Jiaqi Liu, Jinbao Wang, Ziqi Luo, Jinfan Wang, and Feng Zheng.
\newblock Easynet: An easy network for 3d industrial anomaly detection.
\newblock In {\em Proceedings of the 31st ACM International Conference on Multimedia}, pages 7038--7046, 2023.

\bibitem{MulSen-AD}
Wenqiao Li, Bozhong Zheng, Xiaohao Xu, Jinye Gan, Fading Lu, Xiang Li, Na~Ni, Zheng Tian, Xiaonan Huang, Shenghua Gao, et~al.
\newblock Multi-sensor object anomaly detection: Unifying appearance, geometry, and internal properties.
\newblock {\em arXiv preprint arXiv:2412.14592}, 2024.

\bibitem{liu2024learning}
Chieh Liu, Yu-Min Chu, Ting-I Hsieh, Hwann-Tzong Chen, and Tyng-Luh Liu.
\newblock Learning diffusion models for multi-view anomaly detection.
\newblock In {\em European Conference on Computer Vision}, pages 328--345. Springer, 2024.

\bibitem{zhang2024attention}
Yiheng Zhang, Yunkang Cao, Tianhang Zhang, and Weiming Shen.
\newblock Attention fusion reverse distillation for multi-lighting image anomaly detection.
\newblock In {\em 2024 IEEE 20th International Conference on Automation Science and Engineering (CASE)}, pages 2134--2139. IEEE, 2024.

\bibitem{he2024learning}
Haoyang He, Jiangning Zhang, Guanzhong Tian, Chengjie Wang, and Lei Xie.
\newblock Learning multi-view anomaly detection.
\newblock {\em arXiv preprint arXiv:2407.11935}, 2024.

\bibitem{kruse2024splatpose}
Mathis Kruse, Marco Rudolph, Dominik Woiwode, and Bodo Rosenhahn.
\newblock Splatpose \& detect: Pose-agnostic 3d anomaly detection.
\newblock In {\em Proceedings of the IEEE/CVF Conference on Computer Vision and Pattern Recognition}, pages 3950--3960, 2024.

\bibitem{RD++}
Tran~Dinh Tien, Anh~Tuan Nguyen, Nguyen~Hoang Tran, Ta~Duc Huy, Soan Duong, Chanh D~Tr Nguyen, and Steven~QH Truong.
\newblock Revisiting reverse distillation for anomaly detection.
\newblock In {\em Proceedings of the IEEE/CVF Conference on Computer Vision and Pattern Recognition}, pages 24511--24520, 2023.
\newblock doi: {10.1109/CVPR52729.2023.02348}.

\bibitem{zhou2024msflow}
Yixuan Zhou, Xing Xu, Jingkuan Song, Fumin Shen, and Heng~Tao Shen.
\newblock Msflow: Multiscale flow-based framework for unsupervised anomaly detection.
\newblock {\em IEEE Transactions on Neural Networks and Learning Systems}, 2024.

\bibitem{VarAD}
Yunkang Cao, Haiming Yao, Wei Luo, and Weiming Shen.
\newblock Varad: Lightweight high-resolution image anomaly detection via visual autoregressive modeling.
\newblock {\em IEEE Transactions on Industrial Informatics}, 21(4):3246--3255, 2025.

\end{thebibliography}

}

%%%%%%%%%%%%%%%%%%%%%%%%%%%%%%%%%%%%%%%%%%%%%%%%%%%%%%%%%%%%
\clearpage
\appendix
% \title{Supplementary Material for ``Visual Anomaly Detection under Complex View-Illumination Interplay:  A Large-Scale Benchmark''}

\section{Appendix}
The supplementary material includes the following sections to provide additional support for the main manuscript:

\begin{itemize}[label=---, left=1.5em]
    \item \textbf{Sec.~\ref{supp:object_selection}}: Details about our object selection protocol.
    \item \textbf{Sec.~\ref{supp:collection_details}}: More data collection details for \dataset{}.
    \item \textbf{Sec.~\ref{supp:anomaly_visualization}}: Anomaly detection visualizations on \dataset{} for more methods.
    \item \textbf{Sec.~\ref{supp:am_multi_illumination}}: Anomaly detection visualizations on \dataset{} for multi-illumination images.
    \item \textbf{Sec.~\ref{supp:dataset_samples}}: More \dataset{} specimen visualizations.
\end{itemize}

\subsection{Object Selection Protocol}~\label{supp:object_selection}
\begin{figure*}[t]
\centering\includegraphics[width=\linewidth]{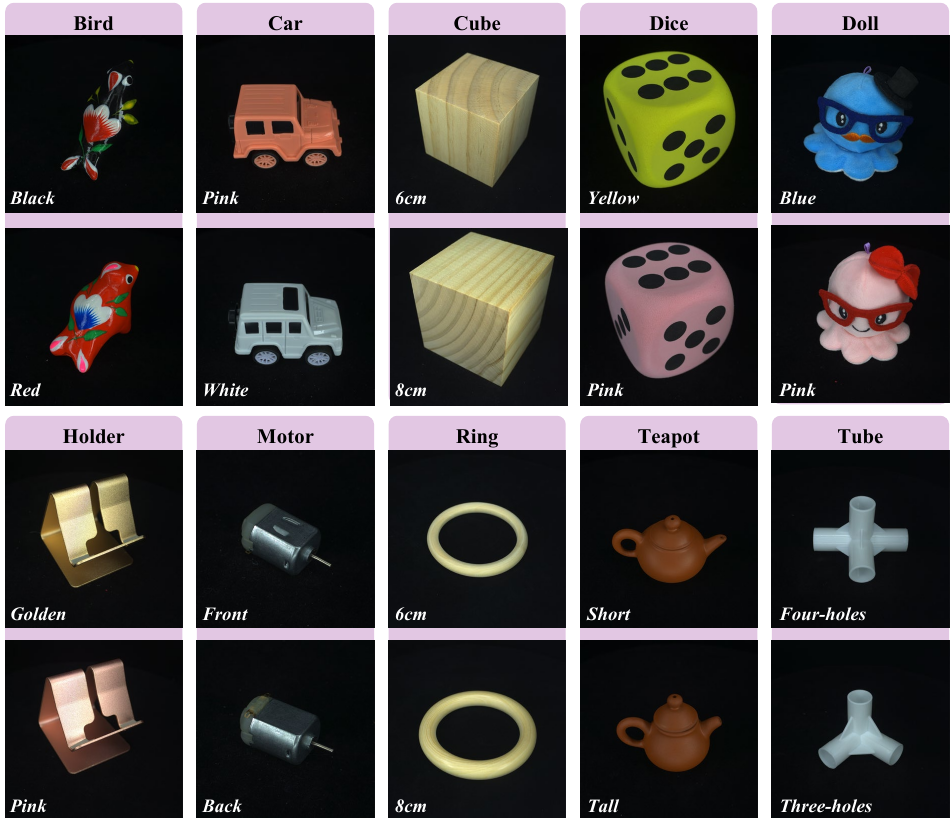}
\caption{\textbf{Visualization of all the categories in \dataset{}}. Each group presents dual sub-categories.
}
% \vspace{-3mm}
\label{fig:sub_cate}
\end{figure*}

To establish a comprehensive anomaly detection dataset that balances ecological validity and methodological challenge, our object selection protocol adheres to three fundamental criteria: (1) \textit{Material Diversity}, ensuring representation of distinct physical properties including clay, plastic, wood, fabric, and metal substrates; (2) \textit{Shape Complexity}, prioritizing objects with intricate three-dimensional geometries or high surface detail density; and (3) \textit{Application Representativeness}, focusing on artifacts prevalent in industrial manufacturing contexts and domestic environments to maximize practical relevance.
\noindent Guided by these principles, we curated ten object categories spanning multiple material domains: Bird, Car, Cube, Dice, Doll, Holder, Motor, Ring, Teapot, and Tube. To amplify dataset versatility and facilitate research in generalized VAD~\cite{AdaCLIP}, each category contains two distinct sub-categories exhibiting systematic variations, as visualized in Fig.~\ref{fig:sub_cate}. These subtype differentiations manifest through either chromatic dissimilarity (\eg, ``Black Bird'' versus ``Red Bird'') or geometric disparity (\eg, ``Tall Teapot'' versus ``Short Teapot''). Comprehensive categorical specifications, including dimensional parameters and material compositions, are tabulated in Table~\ref{tab:M2AD_detailed_info}. The multi-faceted differentiation strategy implemented in \dataset{} ensures both intra-class variance for robustness testing and inter-class diversity for cross-domain generalization analysis.

% \subsection{Object Selection Protocol}
% To construct an anomaly detection dataset with wide adaptability and challenges, we followed the following principles in detected object selection: (1) Material diversity:  Different physical properties need to be covered such as clay, plastic, wood, fabric, and metal; (2) High shape complexity: Objects with three-dimensional structures or rich details should be given priority. (3) Application representativeness: Select common targets in industrial manufacturing or daily scenarios to ensure that the research has practical application value. Therefore, we select ten object categories: Bird, Car, Cube, Dice, Doll, Holder, Motor, Ring, Teapot, and Tube across wide materials. To further enhance diversity and encourage further research on generalized VAD~\cite{AdaCLIP}, we have selected two subtypes for each category, as shown in Fig.~\ref{fig:sub_cate}. The differences between the subtypes can be color, such as ``Black Bird" and ``Red Bird'', or shape/size, such as ``Tall Teapot'' and ``Short Teapot''. The details are shown in Table \ref{tab:M2AD_detailed_info}. Fig.~\ref{fig:sub_cate} visualizes all the sub-categories in \dataset{}.

% \usepackage{booktabs}
\begin{table}[h]
\centering
\caption{Details about the materials and sub-category characteristics of \dataset{}.}
\label{tab:M2AD_detailed_info}
\resizebox{\linewidth}{!}{
\setlength\tabcolsep{36.0pt}
\begin{tabular}{lcc}
\toprule[1.5pt]
\textbf{Category} &
  \textbf{Sub-category} &
  \textbf{Material} \\ \midrule  
Bird & Black / Red & Clay \\
Car & Pink / White & Plastic \\
Cube & 6cm / 8cm   & Wood \\
Dice & Yellow / Pink  & Fabric \\
Doll & Blue / Pink  & Fabric \\
Holder & Golden / Pink & Metal \\
Motor & Front / Back & Metal \\
Ring & 6cm / 8cm  & Wood \\
Teapot & Short / Tall & Clay \\
Tube & Four-holes / Three-holes & Plastic \\ 
\bottomrule[1.5pt]
\end{tabular}
}
\end{table}

\subsection{Data Collection Details}~\label{supp:collection_details}

\begin{figure*}[t]
\centering\includegraphics[width=\linewidth]{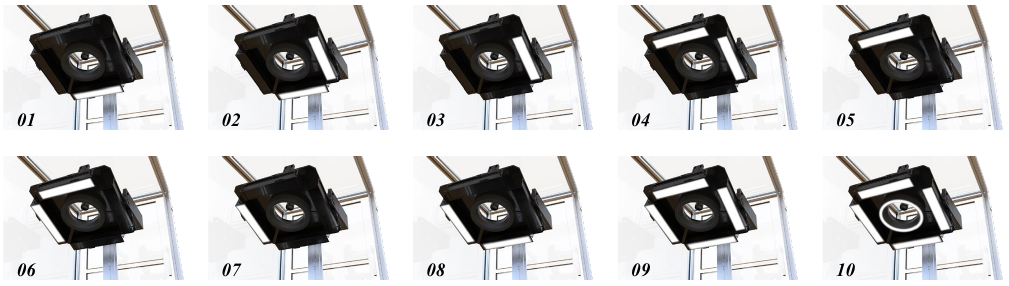}
\caption{\textbf{Schematic illustrations of distinct illumination configurations}. Through programmable control of the photometric illumination module, we sequentially generate ten distinct illumination conditions and acquire corresponding multi-illumination image sequences for comprehensive analysis.
}
% \vspace{-3mm}
\label{fig:illumination_supp}
\end{figure*}

The \dataset{} dataset was acquired through a systematic imaging protocol employing our configurable imaging prototype. The acquisition framework utilizes two principal components: (1) a high-precision motorized turntable with \SI{\pm 0.5}{\degree} angular repeatability for viewpoint control, and (2) a programmable photometric illumination module with configurable source combinations. 
To ensure comprehensive spatial sampling, the rotational stage was programmed to increment in \SI{30}{\degree} angular steps, yielding 12 distinct viewing perspectives per full rotation. At each angular position, the illumination system sequentially activated ten spectrally-tuned light source configurations, as visualized in Fig.~\ref{fig:illumination_supp}. This sampling strategy produces $12 \times 10 = 120$ unique image captures per specimen.

% Affected by the detection efficiency, the number of views and illuminations obtained in the real scene is usually limited. Therefore, we randomly select fewer images from the images of each sample for testing to explore the impact of different numbers of views and illuminations on performance. This exploration will encourage a more effective utilization on multi-views and multi-illuminations data. Specifically, in the experiment in \textbf{Fig. 4(a) from the main text}, we randomly selected 24, 48, 72 and 96 images from 120 images of a sample  to explore the object-wise performance changes. In the experiment in \textbf{Fig. 4(b)}, we fixed 2, 4, 6 and 8 illuminations from all illuminations to explore the variations in image-level and pixel-level performance.
% \noindent\textbf{Number of Images} XXX

% \noindent\textbf{Number of Illuminations} XXX

\begin{figure*}[t]
\centering\includegraphics[width=\linewidth]{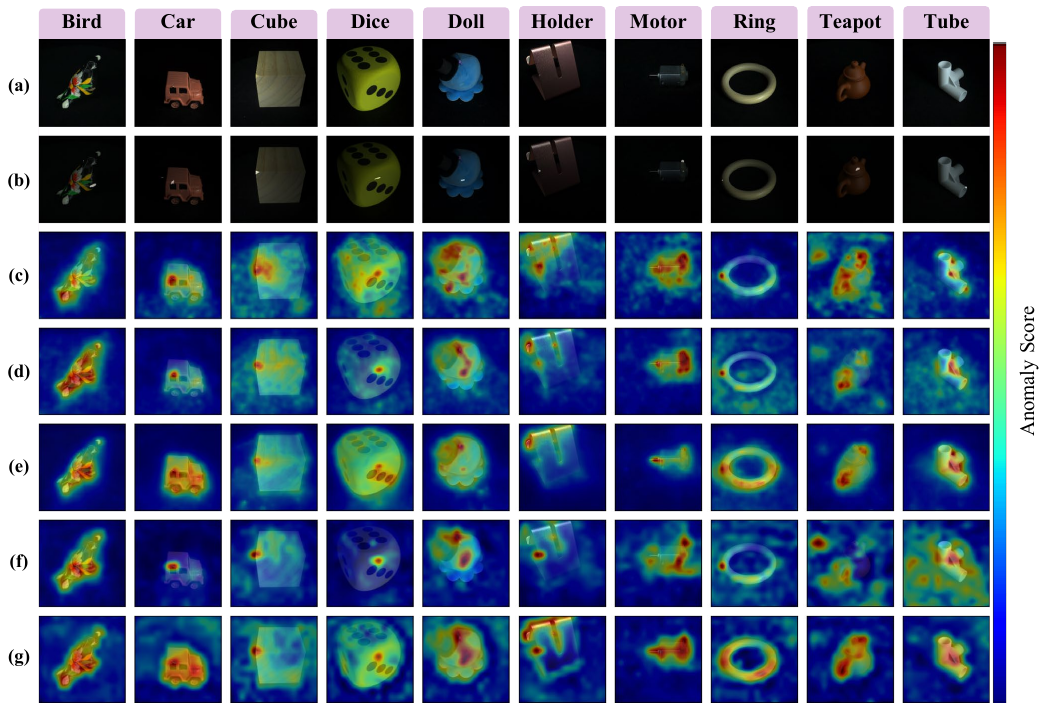}
\caption{\textbf{Visualization of anomaly detection results}. \textbf{(a)} Input image, \textbf{(b)} ground truth (anomalies highlighted in white), \textbf{(c)$\sim$(g)} predicted anomaly maps by CDO~\cite{CDO}, RD++~\cite{RD++}, MSFlow~\cite{zhou2024msflow} Dinomaly~\cite{dinomaly}, and INP-Former~\cite{INP-Former}, respectively. Zoom in for a clearer view.    
}
% \vspace{-3mm}
\label{fig:am_visualizations_supp}
\end{figure*}
\begin{figure*}[t]
\centering\includegraphics[width=\linewidth]{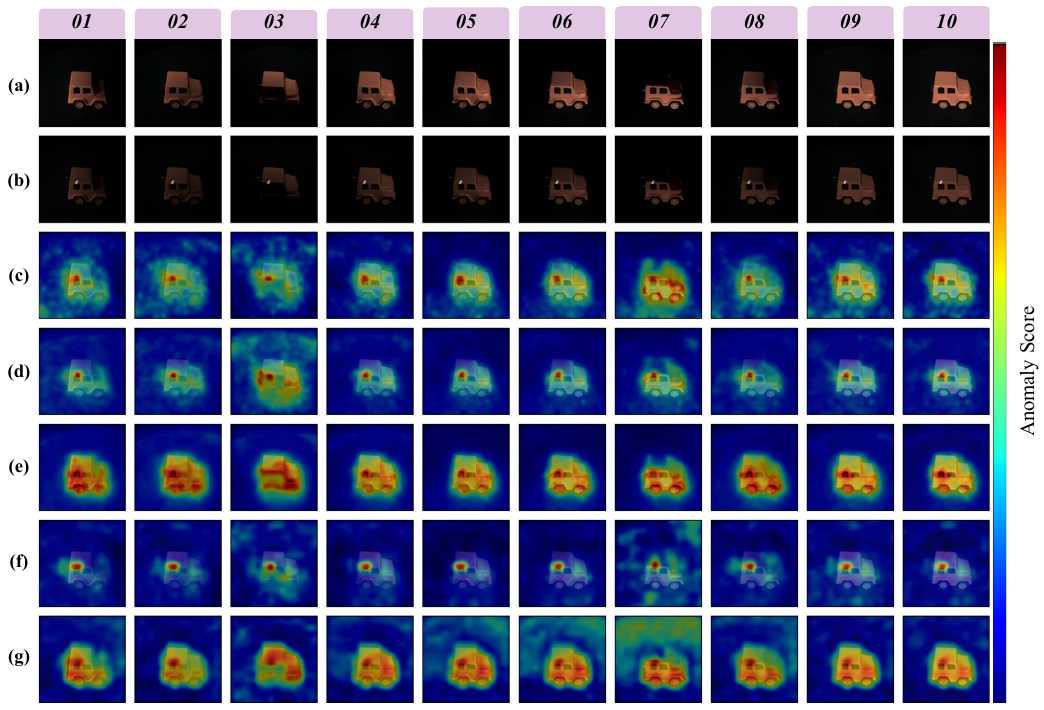}
\caption{\textbf{Visualization of anomaly detection results for multi-illumination images}. 01$\sim$10 corresponds to the illumination conditions in Fig.~\ref{fig:illumination_supp}. \textbf{(a)} Input image, \textbf{(b)} ground truth (anomalies highlighted in white), \textbf{(c)$\sim$(g)} predicted anomaly maps by CDO~\cite{CDO}, RD++~\cite{RD++}, MSFlow~\cite{zhou2024msflow} Dinomaly~\cite{dinomaly}, and INP-Former~\cite{INP-Former}, respectively. Zoom in for a clearer view.    
}
% \vspace{-3mm}
\label{fig:multi_illum_car}
\end{figure*}

\begin{figure*}[t]
\centering\includegraphics[width=\linewidth]{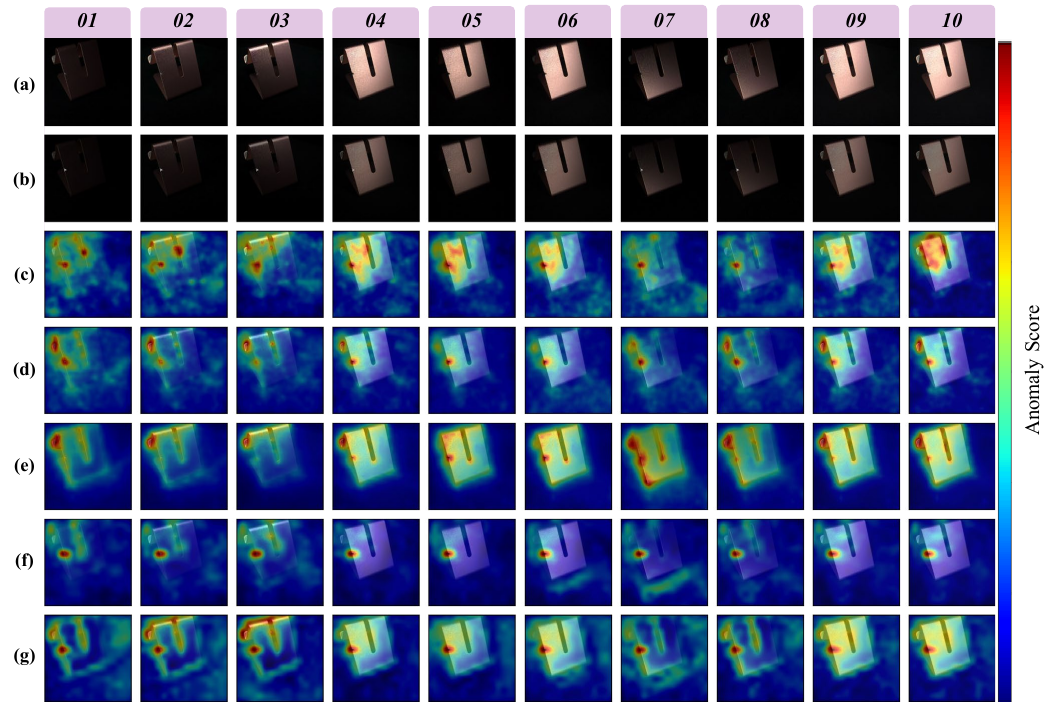}
\caption{\textbf{Visualization of anomaly detection results for multi-illumination images}. 01$\sim$10 corresponds to the illumination conditions in Fig.~\ref{fig:illumination_supp}. \textbf{(a)} Input image, \textbf{(b)} ground truth (anomalies highlighted in white), \textbf{(c)$\sim$(g)} predicted anomaly maps by CDO~\cite{CDO}, RD++~\cite{RD++}, MSFlow~\cite{zhou2024msflow} Dinomaly~\cite{dinomaly}, and INP-Former~\cite{INP-Former}, respectively. Zoom in for a clearer view.    
}
% \vspace{-3mm}
\label{fig:multi_illum_holder}
\end{figure*}

\subsection{Anomaly Detection Visualizations for More Methods}
\label{supp:anomaly_visualization}

Fig.~\ref{fig:am_visualizations_supp} presents a comparative visualization of predicted anomaly maps generated by various benchmark methods. The qualitative analysis reveals a consistent challenge across all approaches: precise anomaly detection at the individual image level remains elusive, particularly given the presence of subtle anomalies combined with suboptimal imaging conditions. This performance gap underscores two critical research imperatives. First, there exists a pressing need to develop noise-robust computational frameworks capable of addressing the challenges posed by real-world imaging artifacts. Second, significant potential resides in designing methodologies that effectively leverage the sequential information inherent in our \dataset{} architecture, which may substantially improve yield rates in practical applications. The current limitations demonstrated in these visualizations highlight the necessity for fundamental algorithmic innovations rather than incremental improvements to existing paradigms.

\subsection{Anomaly Detection Visualizations for Multi-Illumination Images}\label{supp:am_multi_illumination}

Fig.~\ref{fig:multi_illum_car} and~\ref{fig:multi_illum_holder} demonstrate the comparative performance of selected anomaly detection methods across varying illumination conditions. The results reveal a critical dependency between illumination dynamics and anomaly visibility: while certain lighting configurations enable effective identification of anomalous regions, others introduce substantial environmental interference that obscures detection patterns. This illumination-induced variability manifests as significant noise interference in model predictions, particularly when employing naive aggregation approaches. 

Our benchmark analysis indicates that conventional averaging strategies, which indiscriminately combine predictions across all illumination conditions, fail to mitigate this inherent noise propagation. Rather than enhancing detection fidelity, such simplistic fusion mechanisms result in accumulated artifacts that degrade diagnostic precision. This observation is quantitatively corroborated by our ablation studies in Figure~\ref{fig:ablation}, where merely increasing the cardinality of illumination conditions without sophisticated fusion protocols yields diminishing returns. The empirical evidence strongly suggests that illumination multiplicity alone does not guarantee performance improvements unless coupled with intelligent information integration frameworks. 

These findings underscore the necessity for developing context-aware fusion architectures that can adaptively weight illumination-specific features, suppress extraneous noise components, and synthesize discriminative patterns across heterogeneous lighting environments. Future methodological innovations should prioritize illumination-invariant representation learning coupled with dynamic feature selection mechanisms to fully exploit multi-illumination image ensembles.

\subsection{More \textit{\textbf{\texorpdfstring{M$^2$AD}{M2AD}}} Specimen Visualizations}
\label{supp:dataset_samples}

\begin{figure*}[t]
\centering\includegraphics[width=\linewidth]{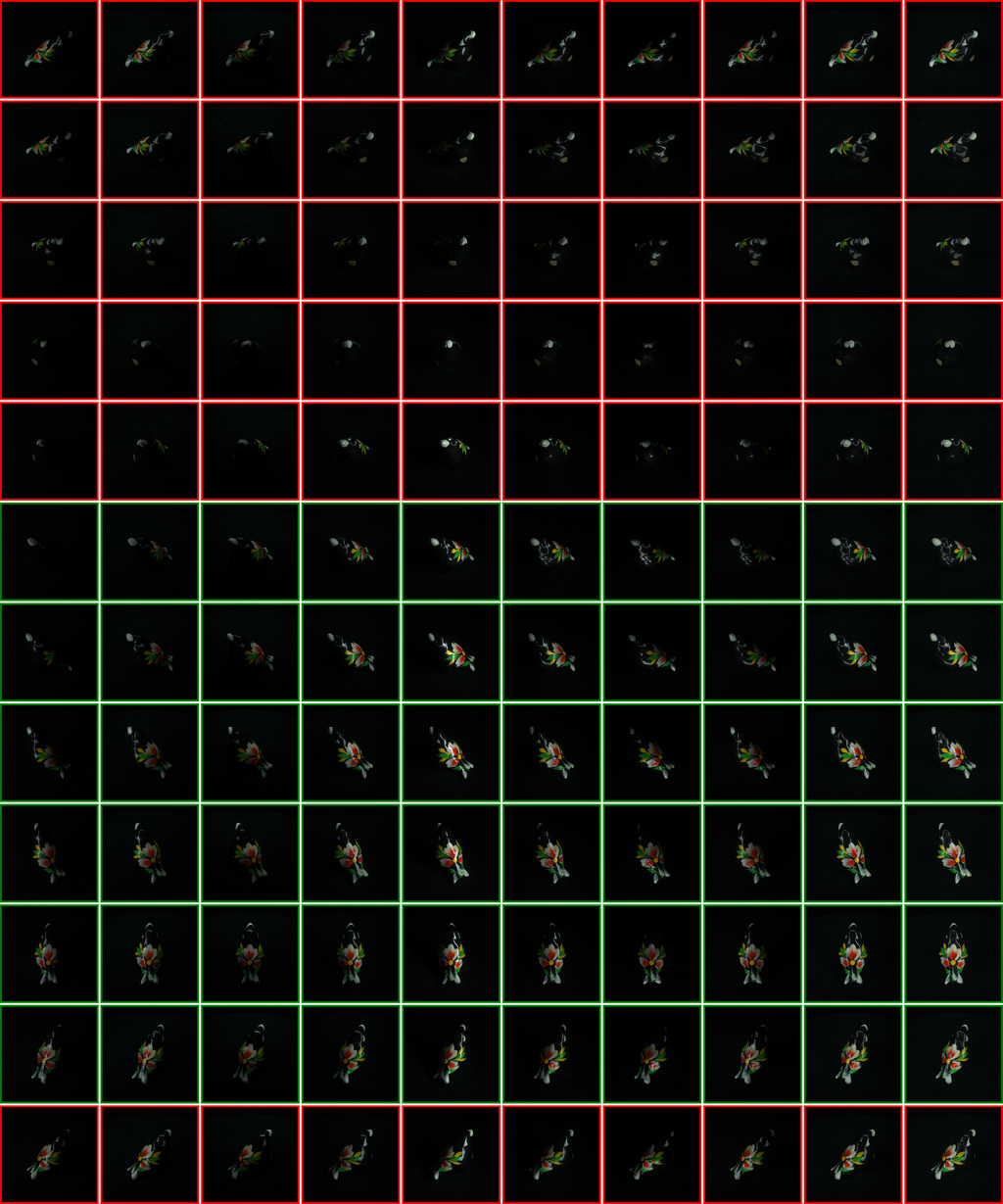}
\caption{\textbf{Visualization of \textit{Bird}}. From top to bottom: images organized by views; from left to right: images organized by illumination condition. Normal images are highlighted with green borders, whereas abnormal images are marked with red borders for comparison.
}
\label{fig:supp_samples_bird}
% \vspace{-3mm}
\end{figure*}

\begin{figure*}[t]
\centering\includegraphics[width=\linewidth]{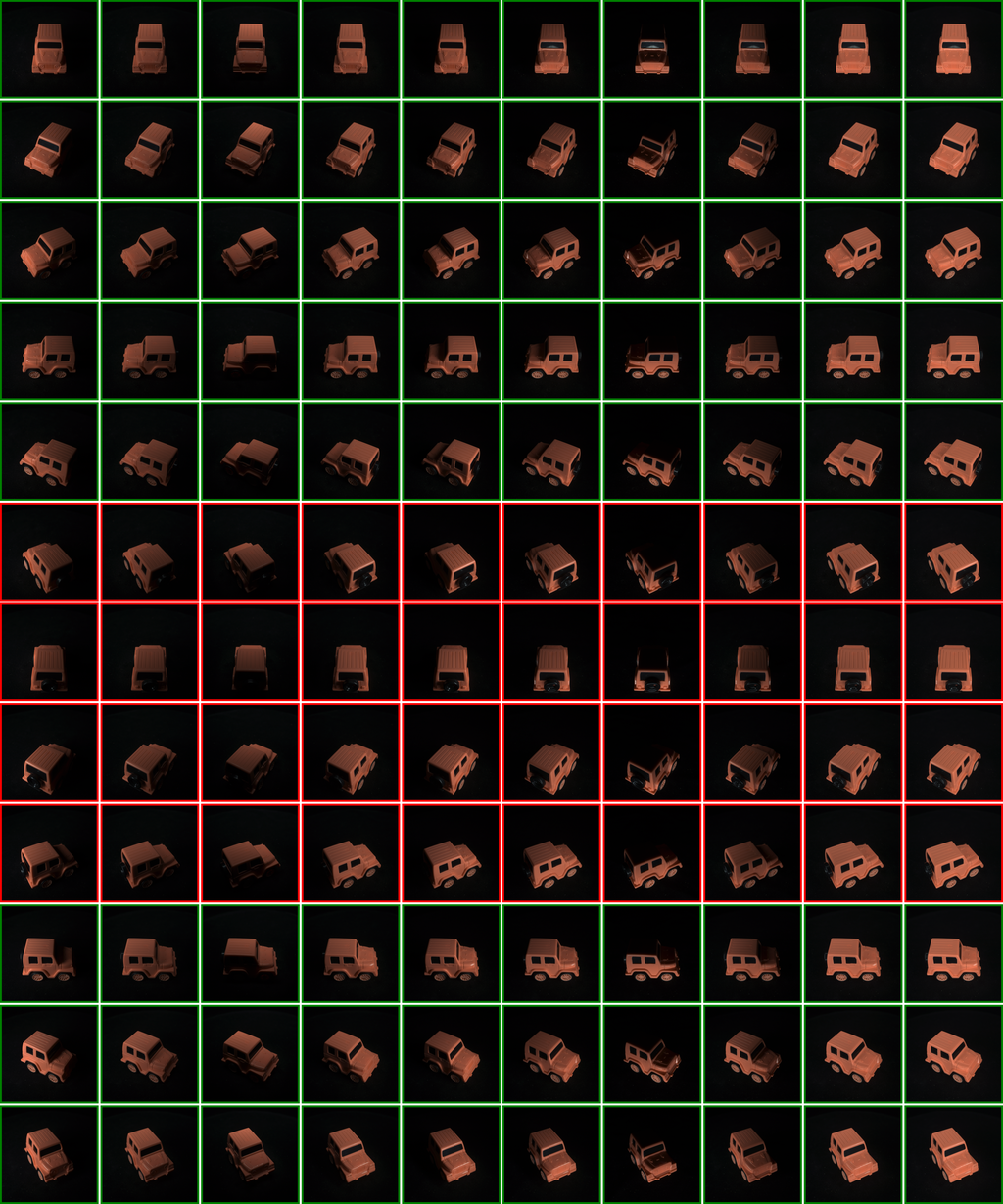}
\caption{\textbf{Visualization of \textit{Car}}. From top to bottom: images organized by views; from left to right: images organized by illumination condition. Normal images are highlighted with green borders, whereas abnormal images are marked with red borders for comparison.
}
% \vspace{-3mm}
\label{fig:supp_samples_car}
\end{figure*}

\begin{figure*}[t]
\centering\includegraphics[width=\linewidth]{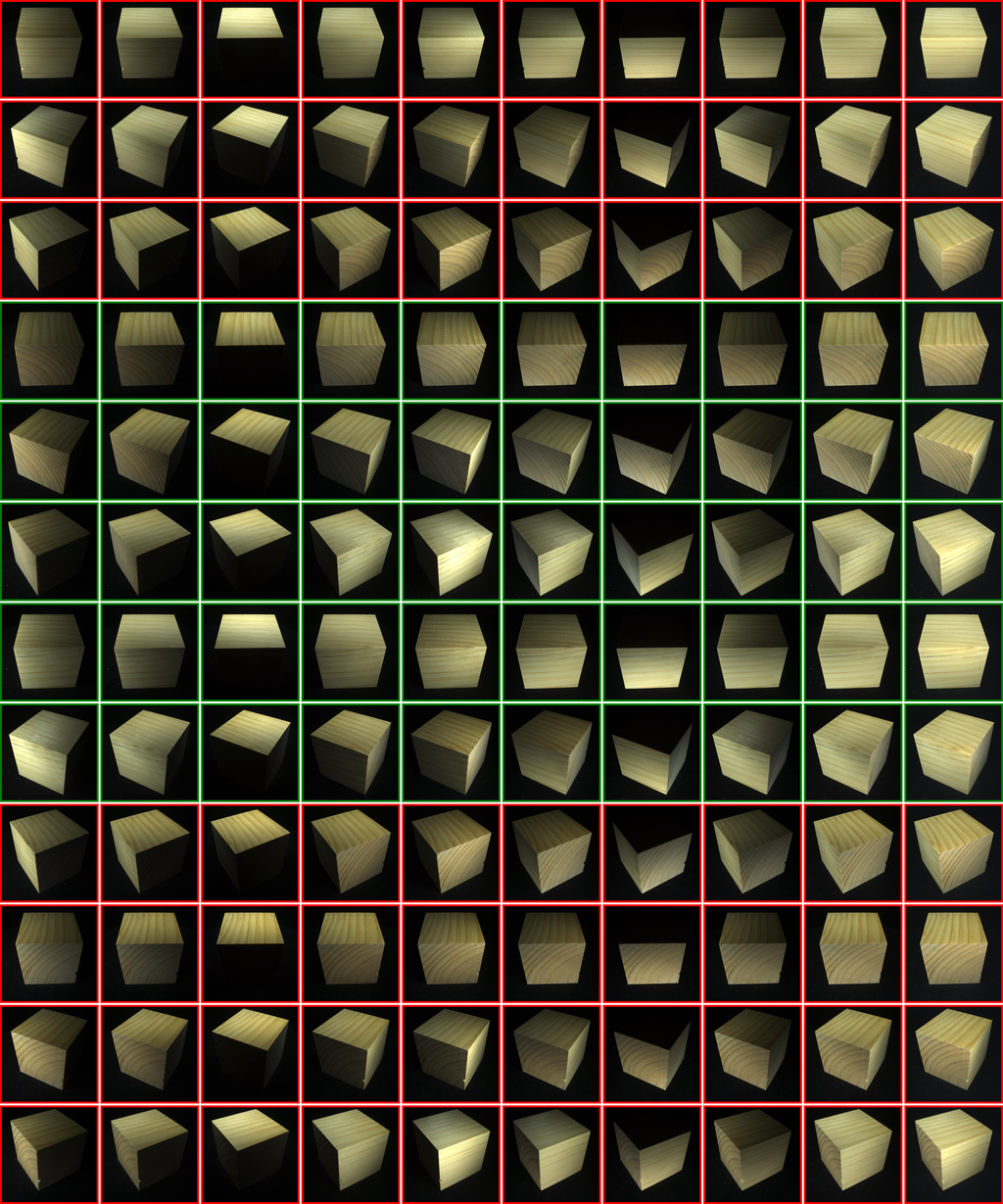}
\caption{\textbf{Visualization of \textit{Cube}}. From top to bottom: images organized by views; from left to right: images organized by illumination condition. Normal images are highlighted with green borders, whereas abnormal images are marked with red borders for comparison.
}
% \vspace{-3mm}
\label{fig:supp_samples_cube}
\end{figure*}

\begin{figure*}[t]
\centering\includegraphics[width=\linewidth]{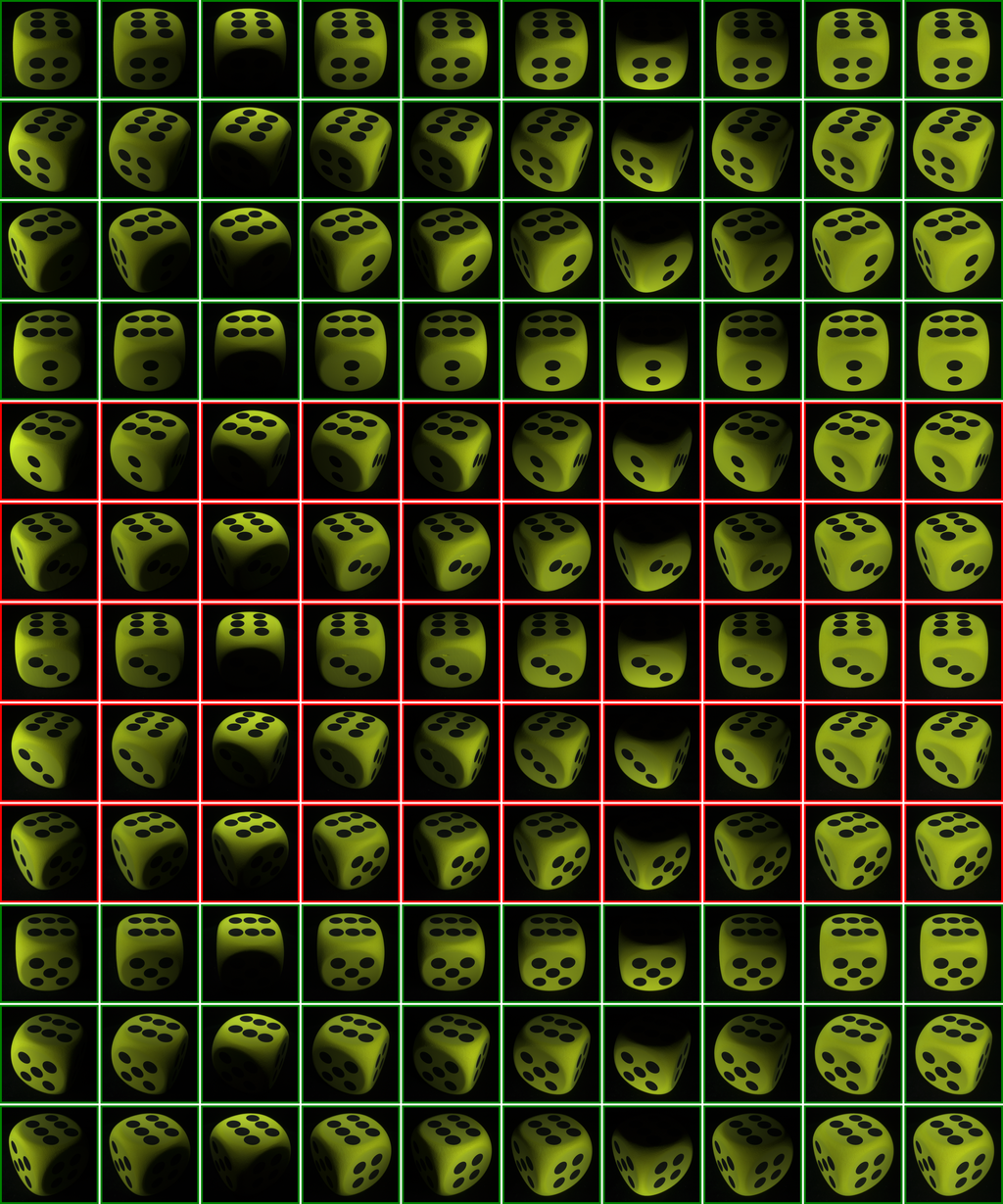}
\caption{\textbf{Visualization of \textit{Dice}}. From top to bottom: images organized by views; from left to right: images organized by illumination condition. Normal images are highlighted with green borders, whereas abnormal images are marked with red borders for comparison.
}
% \vspace{-3mm}
\label{fig:supp_samples_dice}
\end{figure*}

\begin{figure*}[t]
\centering\includegraphics[width=\linewidth]{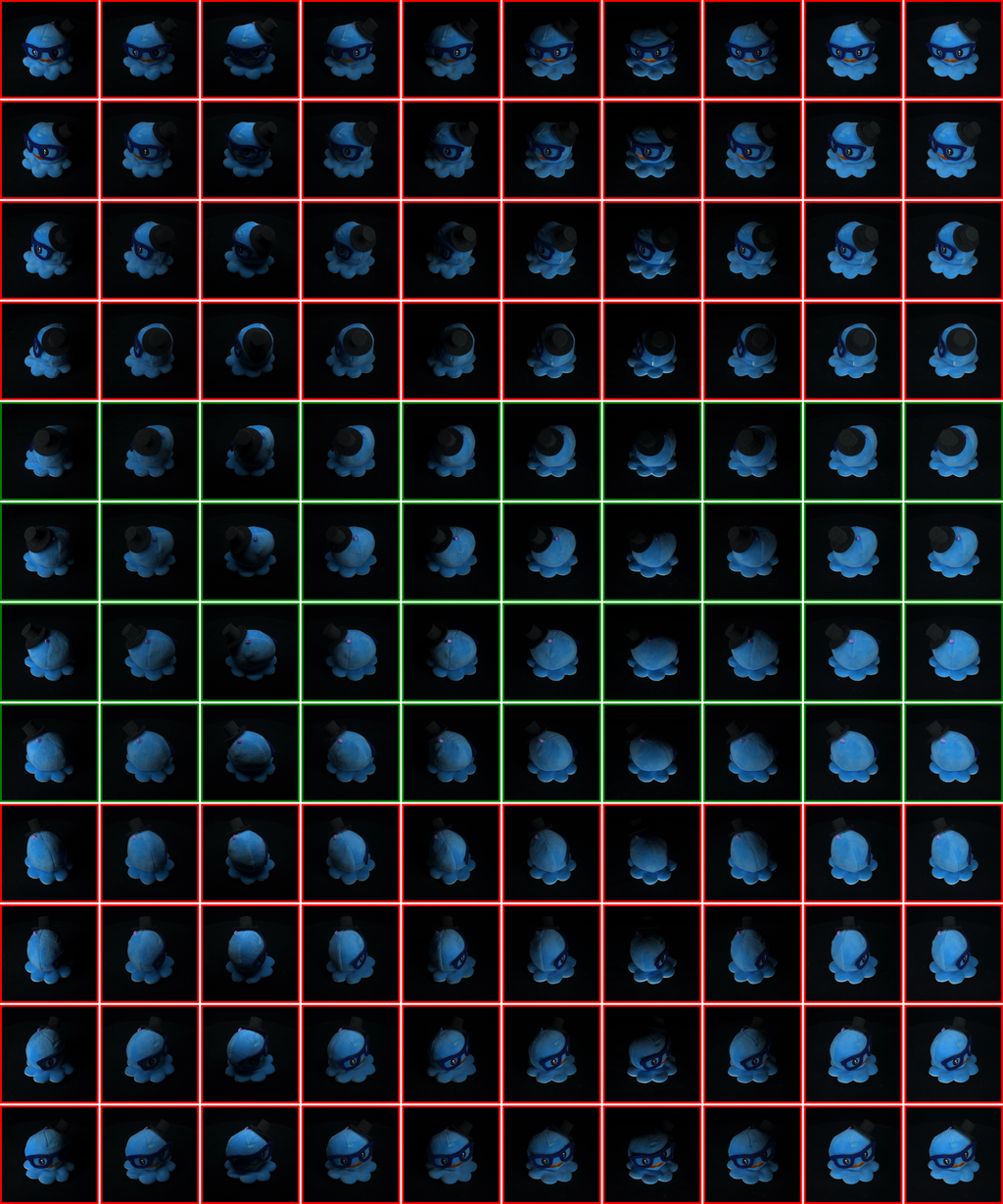}
\caption{\textbf{Visualization of \textit{Doll}}. From top to bottom: images organized by views; from left to right: images organized by illumination condition. Normal images are highlighted with green borders, whereas abnormal images are marked with red borders for comparison.
}
% \vspace{-3mm}
\label{fig:supp_samples_doll}
\end{figure*}

\begin{figure*}[t]
\centering\includegraphics[width=\linewidth]{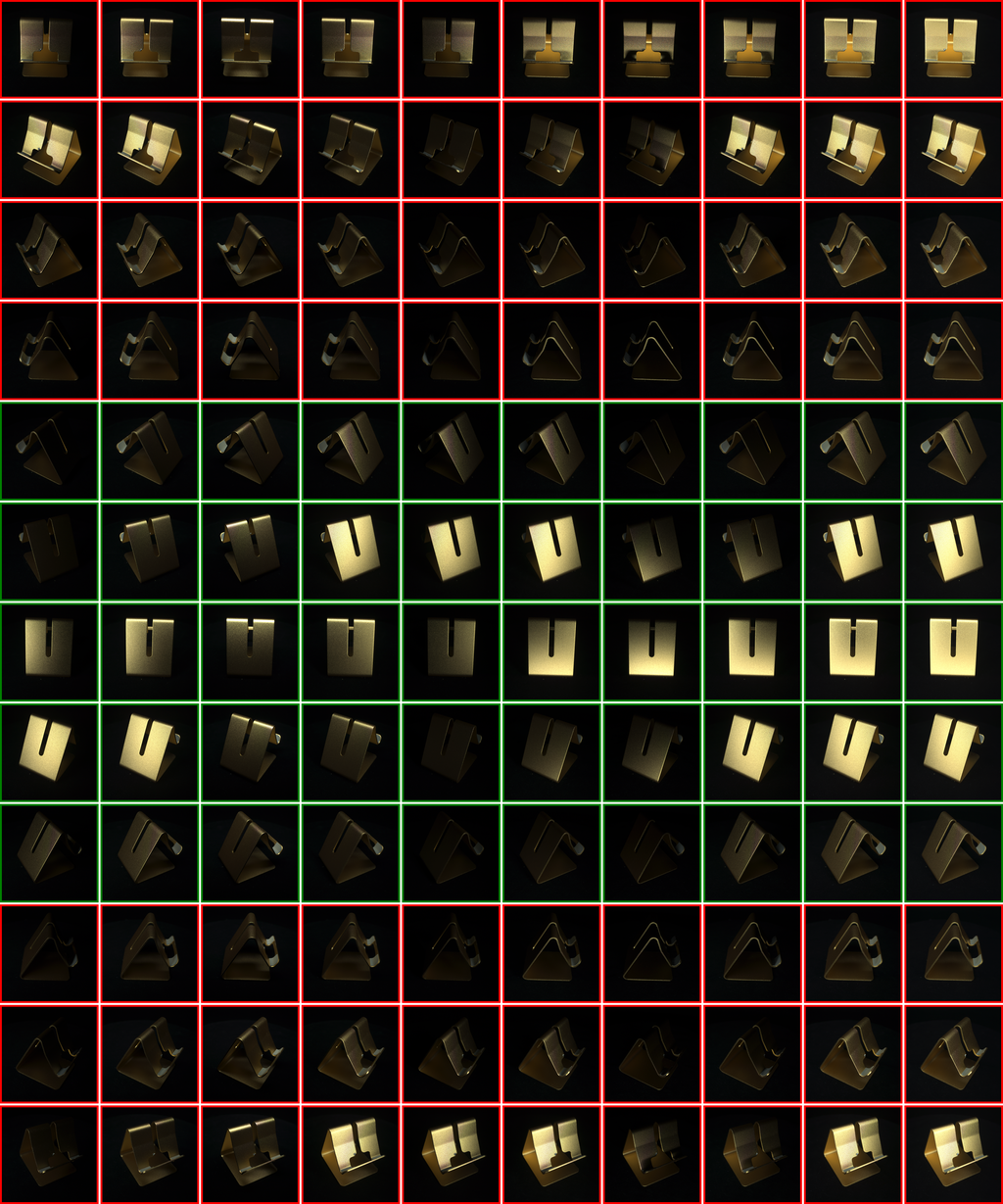}
\caption{\textbf{Visualization of \textit{Holder}}. From top to bottom: images organized by views; from left to right: images organized by illumination condition. Normal images are highlighted with green borders, whereas abnormal images are marked with red borders for comparison.
}
% \vspace{-3mm}
\label{fig:supp_samples_holder}
\end{figure*}

\begin{figure*}[t]
\centering\includegraphics[width=\linewidth]{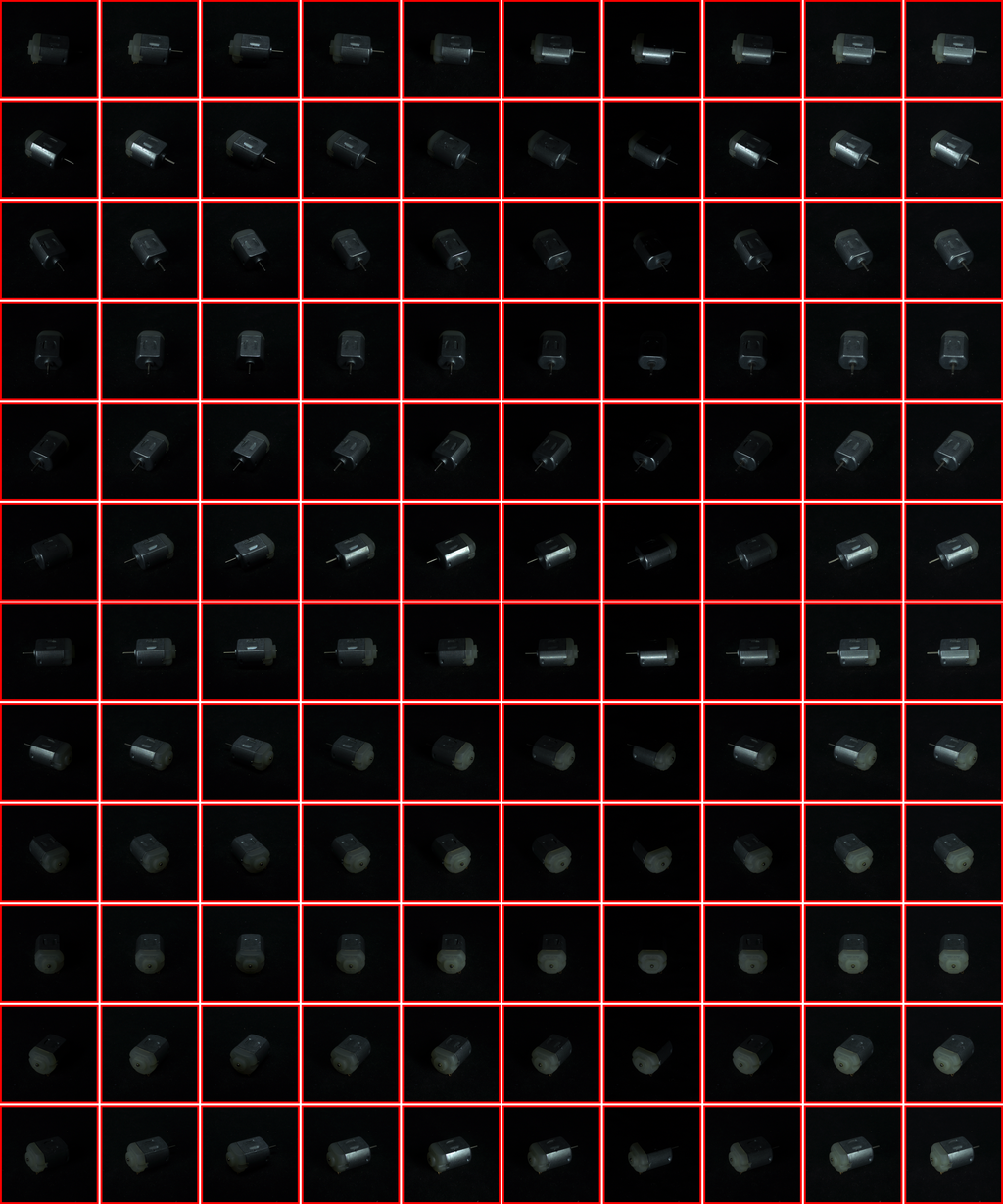}
\caption{\textbf{Visualization of \textit{Motor}}. From top to bottom: images organized by views; from left to right: images organized by illumination condition. Normal images are highlighted with green borders, whereas abnormal images are marked with red borders for comparison.
}
% \vspace{-3mm}
\label{fig:supp_samples_motor}
\end{figure*}

\begin{figure*}[t]
\centering\includegraphics[width=\linewidth]{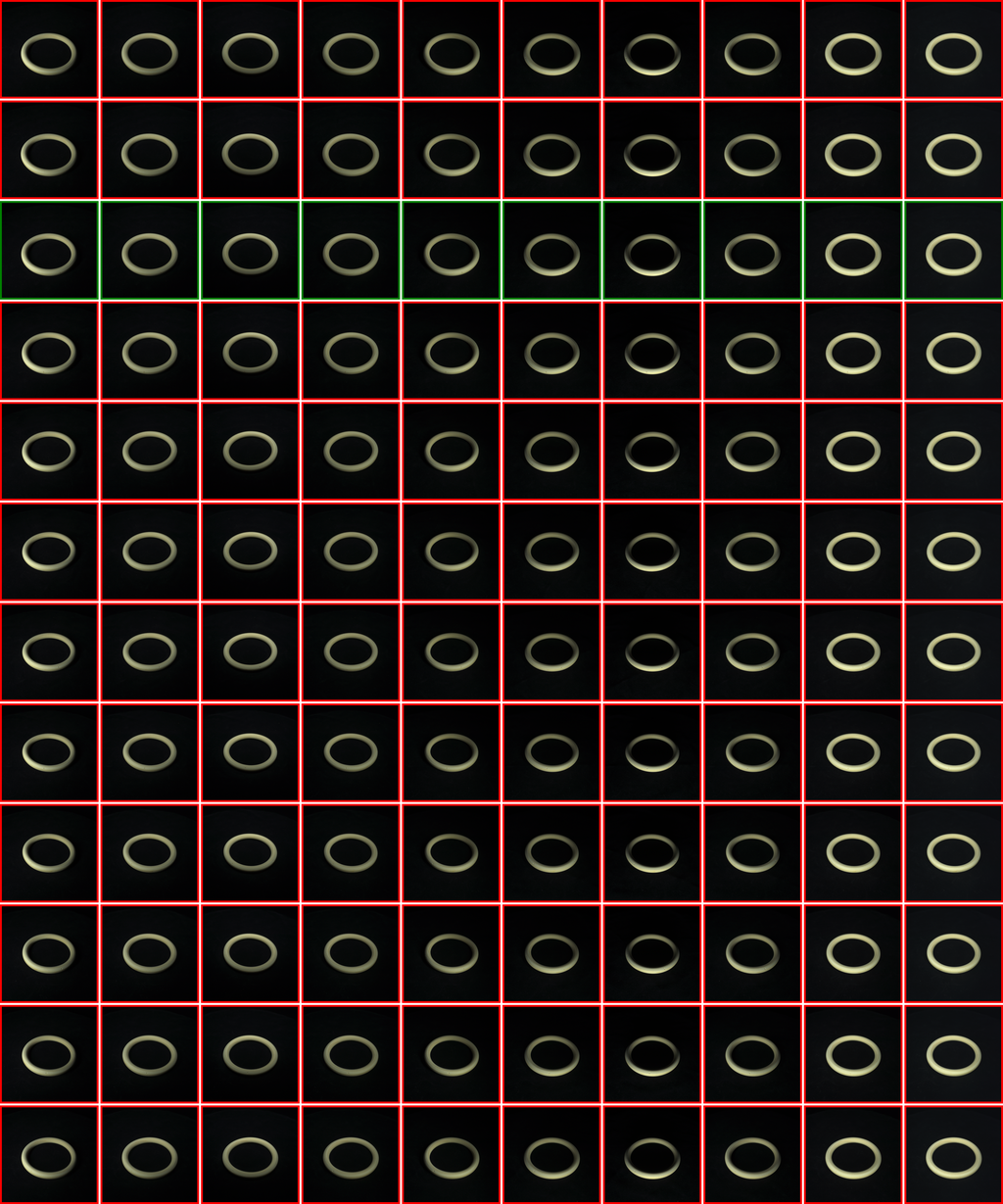}
\caption{\textbf{Visualization of \textit{Ring}}. From top to bottom: images organized by views; from left to right: images organized by illumination condition. Normal images are highlighted with green borders, whereas abnormal images are marked with red borders for comparison.
}
% \vspace{-3mm}
\label{fig:supp_samples_ring}
\end{figure*}

\begin{figure*}[t]
\centering\includegraphics[width=\linewidth]{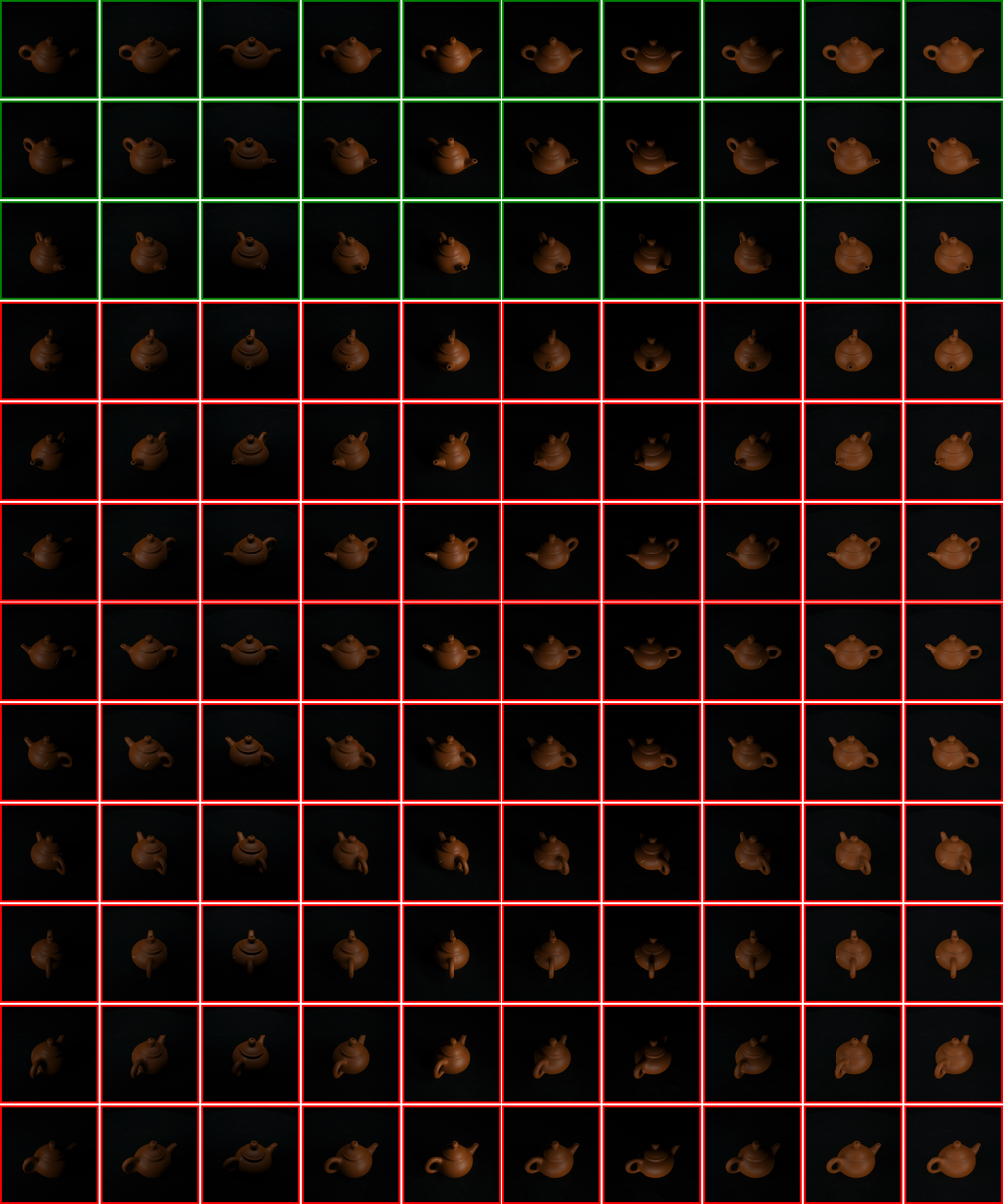}
\caption{\textbf{Visualization of \textit{Teapot}}. From top to bottom: images organized by views; from left to right: images organized by illumination condition. Normal images are highlighted with green borders, whereas abnormal images are marked with red borders for comparison.
}
% \vspace{-3mm}
\label{fig:supp_samples_teapot}
\end{figure*}

\begin{figure*}[t]
\centering\includegraphics[width=\linewidth]{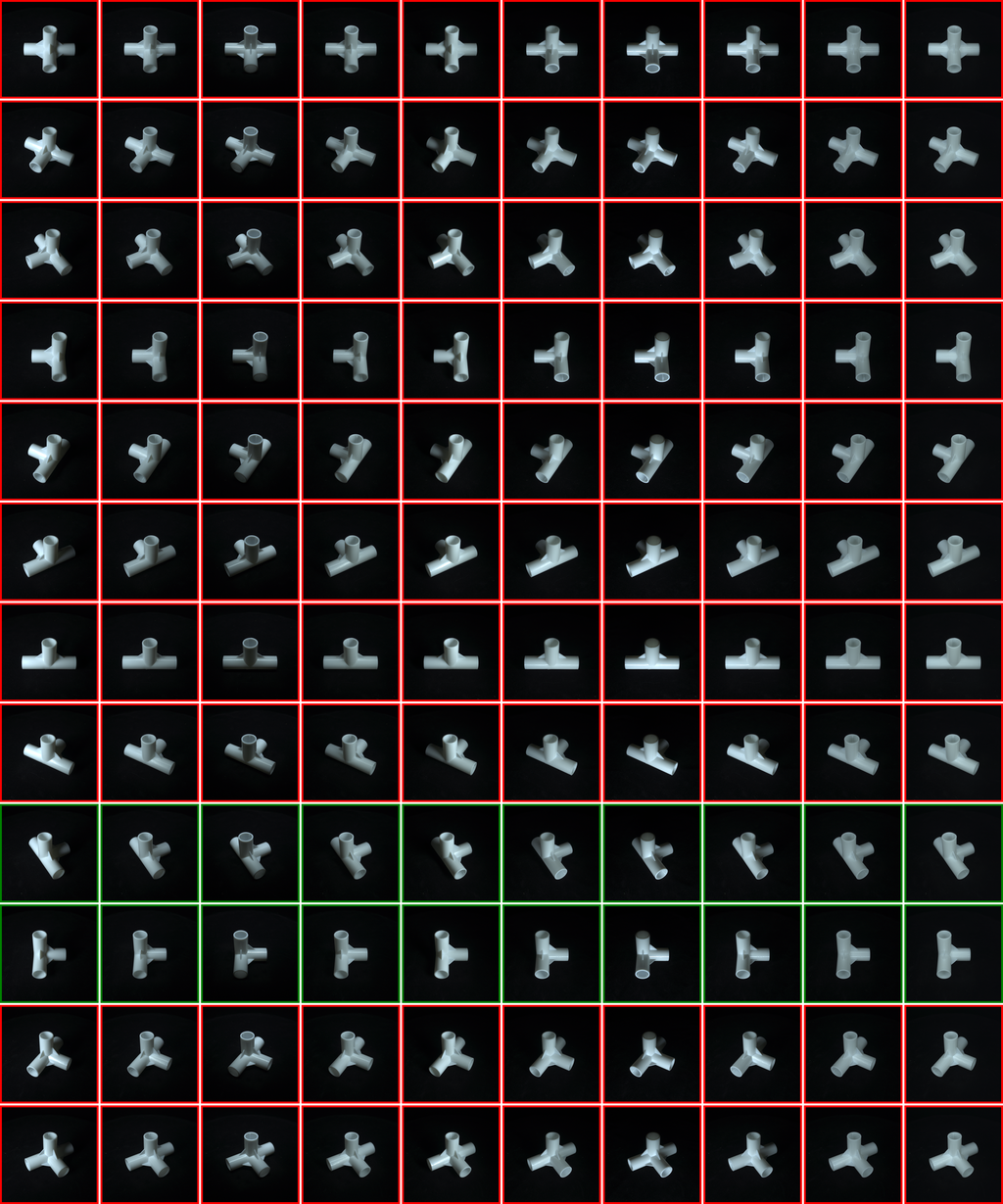}
\caption{\textbf{Visualization of \textit{Tube}}. From top to bottom: images organized by views; from left to right: images organized by illumination condition. Normal images are highlighted with green borders, whereas abnormal images are marked with red borders for comparison.
}
% \vspace{-3mm}
\label{fig:supp_samples_tube}
\end{figure*}

Fig.~\ref{fig:supp_samples_bird}--\ref{fig:supp_samples_tube} systematically present multi-view image sequences (120 frames per specimen) from the \dataset{} collection. These visual sequences exemplify the rich morphological signatures captured through our novel multi-view multi-illumination imaging protocol. The complementary information embedded across different viewing angles and lighting conditions suggests that synergistic integration of these multimodal data streams could substantially enhance performance in visual analysis tasks – a fundamental rationale underlying the \datasetS{} benchmark design.


Notably, our imaging methodology intentionally preserves real-world sensor noise artifacts and photometric variations, including under-exposed regions and specular highlights. This characteristic provides an empirical foundation for evaluating the robustness of VAD algorithms against challenging illumination conditions – a critical requirement for real-world industrial inspection scenarios. The controlled variation in image quality across the dataset enables systematic analysis of failure modes in current computer vision systems.

\end{document}